\documentclass[pdflatex,sn-mathphys]{sn-jnl}
\usepackage{graphics}
\usepackage{subcaption}
\usepackage{quoting}

\jyear{2021}%

\theoremstyle{thmstyleone}%
\theoremstyle{thmstyletwo}%

\theoremstyle{thmstylethree}%

\begin{document}

\title[Article Title]{FedSL: Federated Split Learning on Distributed Sequential Data in Recurrent Neural Networks}

\author*[1]{\fnm{Ali} \sur{Abedi}}\email{ali.abedi@uhn.ca}

\author[1]{\fnm{Shehroz} \sur{S. Khan}}\email{shehroz.khan@uhn.ca}

\affil[1]{\orgdiv{KITE}, \orgname{University Health Network, Canada}}









\abstract{Federated Learning (FL) and Split Learning (SL) are privacy-preserving Machine-Learning (ML) techniques that enable training ML models over data distributed among clients without requiring direct access to their raw data. Existing FL and SL approaches work on horizontally or vertically partitioned data and cannot handle sequentially partitioned data where segments of multiple-segment sequential data are distributed across clients. In this paper, we propose a novel federated split learning framework, FedSL, to train models on distributed sequential data. The most common ML models to train on sequential data are Recurrent Neural Networks (RNNs). Since the proposed framework is privacy preserving, segments of multiple-segment sequential data cannot be shared between clients or between clients and server. To circumvent this limitation, we propose a novel SL approach tailored for RNNs. A RNN is split into sub-networks, and each sub-network is trained on one client containing single segments of multiple-segment training sequences. During local training, the sub-networks on different clients communicate with each other to capture latent dependencies between consecutive segments of multiple-segment sequential data on different clients, but without sharing raw data or complete model parameters. After training local sub-networks with local sequential data segments, all clients send their sub-networks to a federated server where sub-networks are aggregated to generate a global model. The experimental results on simulated and real-world datasets demonstrate that the proposed method successfully trains models on distributed sequential data, while preserving privacy, and outperforms previous FL and centralized learning approaches in terms of achieving higher accuracy in fewer communication rounds.}




\maketitle

\section{Introduction}\label{sec1}

In conventional Machine-Learning (ML) algorithms, models are usually developed in a centralized setting, i.e. the data generated by local clients are firstly aggregated into a server, and the training is performed on the server. However, the client’s personal data is privacy-sensitive and can be large in quantity, which disqualifies centralized ML methods from sending local data to the server. Federated Learning (FL) \cite{mcmahan2017communicationefficient} and Split Learning (SL) \cite{gupta2018distributed} are distributed collaborative ML techniques that decouple model training from the need for direct access to the raw data of clients.

FL \cite{mcmahan2017communicationefficient} enables to collaboratively train a shared model by aggregating locally trained models on clients using their local privacy-sensitive data. On the other hand, SL \cite{gupta2018distributed} trains a model, e.g. a neural network, where the network is split into two sub-networks; one sub-network is trained on client and the other is trained on server. Existing FL and SL approaches are applicable to horizontally \cite{mcmahan2017communicationefficient, he2020fedml, yang2019federated, li2020survey} or vertically \cite{chen2020vafl, cheng2019secureboost, feng2020multiparticipant, liu2020communication} partitioned data, and cannot handle sequentially partitioned data occurring frequently in practice. In sequentially partitioned data, where training sequences are sequentially distributed across clients, consecutive segments of multiple-segment sequential data are generated and available on different clients.

As a concrete example of sequentially partitioned data, suppose one patient is consecutively admitted to two different hospitals each recording sequential vital signs of the patient \cite{pollard2018eicu}, e.g. heart rate. In this case, the heart rate of the patient is sequentially distributed across two hospitals/clients. This multiple-segment sequential data can be used to train ML models to predict, for instance, the mortality of patients. The most common ML models to train on sequential data are Recurrent Neural Networks (RNNs), including the vanilla RNN, Long Short-Term Memory (LSTM), and Gated Recurrent Unit (GRU) \cite{sherstinsky2020fundamentals}. Due to privacy concerns, the clients are not allowed to share the segments of sequential data. One solution to decouple training RNN from segment sharing between clients is to split RNN into sub-networks each being trained on single clients containing single segments of sequential data. Existing SL approaches work on feed-forward neural networks, and cannot be used for RNNs \cite{abuadbba2020use}. RNNs, in order to train on sequentially partitioned data, should be split in a way that preserve latent dependencies between segments of multiple-segment sequential data available on different clients.

In this paper, we present FedSL, a novel collaborative ML framework utilizing FL, along with a novel SL approach tailored for RNNs, to take advantage of both approaches and to train models on sequentially partitioned data, while preserving the privacy of clients' local data. Due to the privacy-preserving property of FL and SL approaches, data sharing between clients and also between clients and server is not allowed. Therefore, in FedSL, model training is performed locally on each client in collaboration with other clients and with the server. SL is performed between clients, and FL is performed between clients and the server. A RNN is split into two sub-networks to be trained with sequential training samples with multiple segments available on different clients. The first sub-network is trained with the first segment in one client, and the second sub-network is trained using the second segment in another client. During SL, these sub-networks of the split RNN, on different clients, collaborate and communicate to learn from their local segments without sharing data or model parameters to other clients. These rounds of SL are repeated whenever new training sequential segments are generated in consecutive clients. After each round of SL, all clients send their locally trained sub-networks to the server where all received sub-networks are aggregated. These aggregated sub-networks are used as the initial model for the next rounds of FedSL. This process continues until the server acknowledges the convergence. Our primary contributions in this paper are summarized as follows:
\begin{itemize}
\item We are first to define and address the problem of training models in a federated setting with multiple-segment sequential data distributed across clients. This data distribution is different from vertical and horizontal partitioned data \cite{yang2019federated}, and none of the previously presented FL or SL approaches have studied this data-partitioning structure \cite{yin2021comprehensive, thapa2021advancements, he2020fedml, yang2019federated, li2020survey, li2020federated}.
\item To address the above problem, we propose a novel architecture that integrates FL and SL. In this architecture, none of the \emph{data}, or \emph{label}, or \emph{complete model parameters} are shared between clients or between clients and server.
\item We propose a novel SL approach for RNNs. Consecutive splits of RNNs are trained on consecutive clients containing consecutive segments of multiple-segment sequential data distributed across clients.
\item We conduct extensive experiments on simulated and real-world sequential multiple-segment datasets, showing that our proposed framework successfully trains sequential models, while preserving privacy, and outperforms previous FL and centralized learning approaches in terms of achieving higher accuracy in fewer communication rounds.
\end{itemize}
The paper is structured as follows. Section \ref{sec:blr} introduces background and related works on FL and SL. In Section \ref{sec:fedsl}, the proposed pathway for analyzing distributed sequential data through splitting RNNs is presented. Section \ref{sec:er} describes experimental settings and results on the proposed methodology. In the end, Section \ref{sec:cf} presents our conclusions and directions for future works.

\section{Background and Literature Review}  \label{sec:blr}
In this Section, after briefly explaining FL and SL, we discuss their related work, and various ways these two frameworks are integrated in previous works.

\subsection{Federated Learning} \label{sec:fl}

FL is a collaborative distributed learning paradigm firstly developed by Google \cite{mcmahan2017communicationefficient} aiming to train ML models on distributed devices having locally generated privacy-sensitive training samples. At the starting round of the FL ($t=0$), the server initializes a global model $W_0$, and sends it to selected clients participating in the current round of FL. After receiving the initial model, each client starts training, updating the initial model, using its local training samples. Then, each client sends the updated model back to the server. The server aggregates all the received models to generate the updated version of the global model, $W_{t+1}$. These rounds of computation-communication continue until the server acknowledges the global model to be converged.

From data-distribution point of view, existing FL approaches are categorized into vertical and horizontal, \cite{yin2021comprehensive, he2020fedml, li2020survey}. The horizontal FL approaches focus on analyzing horizontal data structure in which training samples in different clients are represented by the same features, and training sample IDs in clients are unique, i.e., there are no different versions of one sample ID in different clients. Contrarily, in vertical FL, different clients contain different versions of the same training sample IDs, represented by different features.

Most existing FL approaches are in the category of horizontal FL \cite{mcmahan2017communicationefficient, yang2019federated, li2020survey, he2020fedml}. The vanilla FL (federated averaging, FedAvg) \cite{mcmahan2017communicationefficient}, used for next-word prediction \cite{hard2019federated} and emoji prediction \cite{ramaswamy2019federated} in virtual keyboard in smartphones, is a horizontal FL approach.

Chen et al. \cite{chen2020vafl} proposed a vertical asynchronous FL (VAFL) approach which allows clients to participate in FL asynchronously without coordination with other clients. As it is a privacy-preserving approach, instead of sharing raw data of clients to the server, a local embedding, a mapping reducing the dimensionality of the raw data, is sent to the server, and client’s local model updating is performed with collaboration with the server, independent of other clients. Different classification models were used in VAFL setting, including Logistic Regression (LR), Convolutional Neural Networks (CNNs), and LSTMs on various image and Electronic Health Record (EHR) datasets.

Cheng et al. \cite{cheng2019secureboost} proposed a vertical FL for training gradient boosting decision trees using vertically partitioned training datasets. A privacy-preserving entity alignment method is used to find common users in different clients. They used financial datasets and trained their models in the vertical federated setting to solve binary classification problems.

Other important points about data distribution in FL are the balance of data among clients, and if the data is independent and identically distributed (IID) among clients \cite{mcmahan2017communicationefficient, li2020convergence}. Many works have been done to address the non-IID data issue in FL \cite{he2020fedml}. For instance, Xiang et al. \cite{li2020convergence} proposed an improved version of FedAvg with data sharing in which distributing a small amount of holdout IID data between clients results in a significant improvement in the performance of FL on non-IID data. Briggs et al. \cite{briggs2020federated} proposed to separate clients in clusters based on the similarity of their locally trained models to the global model, and then train clusters of clients independently. In this way, higher accuracy on non-IID data is achieved compared to the vanilla FL.

While most FL frameworks work based on the Stochastic Gradient Descent (SGD) optimization, some other methods proposed modified versions of SGD to improve learning performance. In FedAvg \cite{mcmahan2017communicationefficient}, clients run local SGD for a predetermined number of epochs. A modified version of FedAvg, LoAdaBoost, is presented in \cite{huang2020loadaboost} where for each client, after a certain number of epochs, if the local loss is higher than a threshold, the local epochs of training continue to decrease the local loss, otherwise local training finishes. Another modification of FedAvg is presented by Baheti et al. in \cite{baheti2020federated} in which the weights in FedAvg are modified based on the local loss of clients. The clients with lower local loss will have greater weights in FedAvg. For more information on different FL approaches, see \cite{yin2021comprehensive, thapa2021advancements, he2020fedml, yang2019federated, li2020survey, li2020federated}.

\subsection{Split Learning} \label{sec:sl}
In the vanilla SL algorithm \cite{gupta2018distributed}, from one specific layer, called split layer or cut layer, the neural network is split into two sub-networks. The client, containing training data, performs forward propagation and computes the output of its sub-network up to the cut layer. The cut layer’s output is sent to the server. The server computes the final output, until the last layer of the network. At server’s sub-network, the gradients are backpropagated from the last layer to the cut layer, and the gradient of the cut layer is sent back to the client. The client performs the rest of the backpropagation process from the cut layer to the first layer of the network. This process continues until the client has new training data.

In the above described vanilla SL architecture \cite{gupta2018distributed, thapa2020splitfed, gao2020endtoend, abuadbba2020use}, the server has no direct access to the raw data of clients, and also complete model parameters are not sent to the server. The only information being communicated between clients and the server is the output of the cut layer, from clients to the server, and the gradient of the cut layer, from the server to clients. However, one drawback of the vanilla SL \cite{gupta2018distributed} is that the labels of training samples need to be transmitted from clients to the server. In \cite{vepakomma2018split}, a U-shaped configuration for SL is presented to alleviate the problem of label sharing in SL, where four sub-networks are used to make the model training possible without label sharing.

In Section \ref{sec:fl}, FL algorithms were studied from data-partitioning structure point of view. SL algorithms can also be described based on the structure of data they analyze. The vanilla SL \cite{gupta2018distributed} and U-shaped SL approaches \cite{vepakomma2018split} described above are suitable for horizontal data where training samples in different clients share the same feature space but do not share the same sample ID space. In contrast, vertical SL approaches deal with data structures in which different features of the same training samples are available on different clients. Vepakomma et al. \cite{vepakomma2018split} proposed a vertical SL configuration in which two clients containing different modalities of the same training samples train their specific sub-networks up to the cut layer. Then, the outputs of their sub-networks are concatenated and sent to the server. The server performs forward and backward propagations and sends the gradient to each of the clients to complete the backward propagation and train the overall network.

Gupta and Raskar in \cite{gupta2018distributed} evaluated the performance of horizontal SL with one or more clients collaborating with one server using various image classification datasets. They achieved classification accuracy comparable to the centralized learning using different feed-forward 2D CNNs. According to their experiments, participating more clients in SL causes accuracy to improve significantly. Abuadbba et al. in \cite{abuadbba2020use} examined the application of SL in 1D CNNs for detecting heart abnormalities using medical ECG data and to classify speech signals. They concluded that 1D CNN classifiers under SL are able to achieve the same accuracy as centralized learning.

\subsection{Integrating Federated Learning with Split Learning}  \label{sec:iflsl}
Thapa et al. \cite{thapa2020splitfed} proposed SplitFed framework combining FL and SL to train models on horizontally partitioned data. This framework splits a neural network between each participating client and a \emph{main server}. The clients carry out SL with the main server to train their local models. Then, the clients send their local models to a \emph{fed server} which conducts federated averaging of the received models to create a global model. In this configuration, label sharing is needed, i.e. all clients send the label of their local data to the main server to be able to perform SL.

Han et al. \cite{hanaccelerating} proposed an approach for accelerating FL on horizontally distributed data through integrating FL with SL. In this method, the client-side sub-networks can be updated without receiving the gradient of loss from the server using local-loss-based training.

Mugunthan et al. \cite{mugunthan2021multi} proposed an approach for FL on vertically distributed data, when there are multiple data and label owner clients, through integrating SL with adaptive federated optimizers.

Based on the descriptions of the related works, Table. 1 briefly compares related FL and SL approaches from different points of view. The previous FL \cite{mcmahan2017communicationefficient, chen2020vafl, feng2020multiparticipant, yin2021comprehensive, thapa2021advancements, hanaccelerating} and SL \cite{gupta2018distributed, abuadbba2020use, hanaccelerating, mugunthan2021multi, thapa2021advancements} methods work on horizontally or vertically partitioned data. Our method, FedSL (described in Section \ref{sec:fedsl}), is the first framework that handles sequentially partitioned data in a privacy-preserving manner. Previous SL methods split feed-forward neural networks \cite{gupta2018distributed, abuadbba2020use, thapa2020splitfed, hanaccelerating, mugunthan2021multi, thapa2021advancements}, while FedSL splits recurrent neural networks. In the SplitFed method \cite{thapa2020splitfed}, SL and FL are combined to work on horizontally partitioned data. In SplitFed, after SL between clients and a main server, complete model parameters are sent from clients to the fed server. It is against the main goal of SL which is to avoid complete model sharing among clients and server. In addition, in the SL step of SplitFed, label sharing is required among clients and the split server. However, in FedSL, the complete model is not shared and label sharing is not required between clients, or between clients and server.

\begin{table}[h!]
\centering
\caption{
Comparing FL \cite{mcmahan2017communicationefficient},
VAFL \cite{chen2020vafl}, MMVFL \cite{feng2020multiparticipant},
SL \cite{gupta2018distributed},
1D-CNN SL \cite{abuadbba2020use},
SplitFed \cite{thapa2020splitfed},
Accelerated FL \cite{hanaccelerating},
Multi-VFL \cite{mugunthan2021multi},
and our proposed method, FedSL (described in Section \ref{sec:fedsl}), from different perspectives.}
\resizebox{\columnwidth}{!}{
\begin{tabular}{c c c c c c}
 \hline
 Method & Data & Label & Model & Model & Split \\ [0.5ex]
  & partitioning & sharing & aggregation & split & neural network \\ [0.5ex]
 \hline\hline
 FL \cite{mcmahan2017communicationefficient} & Horizontal & No & Yes & No & - \\ 
 VAFL \cite{chen2020vafl}, MMVFL \cite{feng2020multiparticipant} & Vertical & Yes & Yes & No & - \\
  SL \cite{gupta2018distributed} & Horizontal & Yes & No & Yes & Feed-forward \\
  1D-CNN SL \cite{abuadbba2020use} & Horizontal & Yes & No & Yes & Feed-forward \\
  SplitFed  \cite{thapa2020splitfed} & Horizontal & Yes & Yes & Yes & Feed-forward \\
  
  Accelerated FL  \cite{hanaccelerating} & Horizontal & Yes & Yes & Yes & Feed-forward \\
  
  Multi-VFL  \cite{mugunthan2021multi} & Vertical & Yes & Yes & Yes & Feed-forward \\
  
  \textbf{FedSL (proposed)} & \textbf{Sequential} & \textbf{No} & \textbf{Yes} & \textbf{Yes} & \textbf{Sequential (RNN)} \\
 \hline
\end{tabular}
}
\label{table:1}
\end{table}

\section{FedSL - A Novel Federated Split Learning Framework}  \label{sec:fedsl}

In this Section, after defining the problem of training models on multiple-segment sequential data distributed among clients in Section \ref{sec:pd}, SL in RNNs is introduced in Section \ref{sec:slrnn}, and finally, the overall FedSL framework for analyzing sequentially partitioned data is presented in Section \ref{sec:fedslalgorithm}.

\begin{figure}
\centering
  \includegraphics[scale=.25]{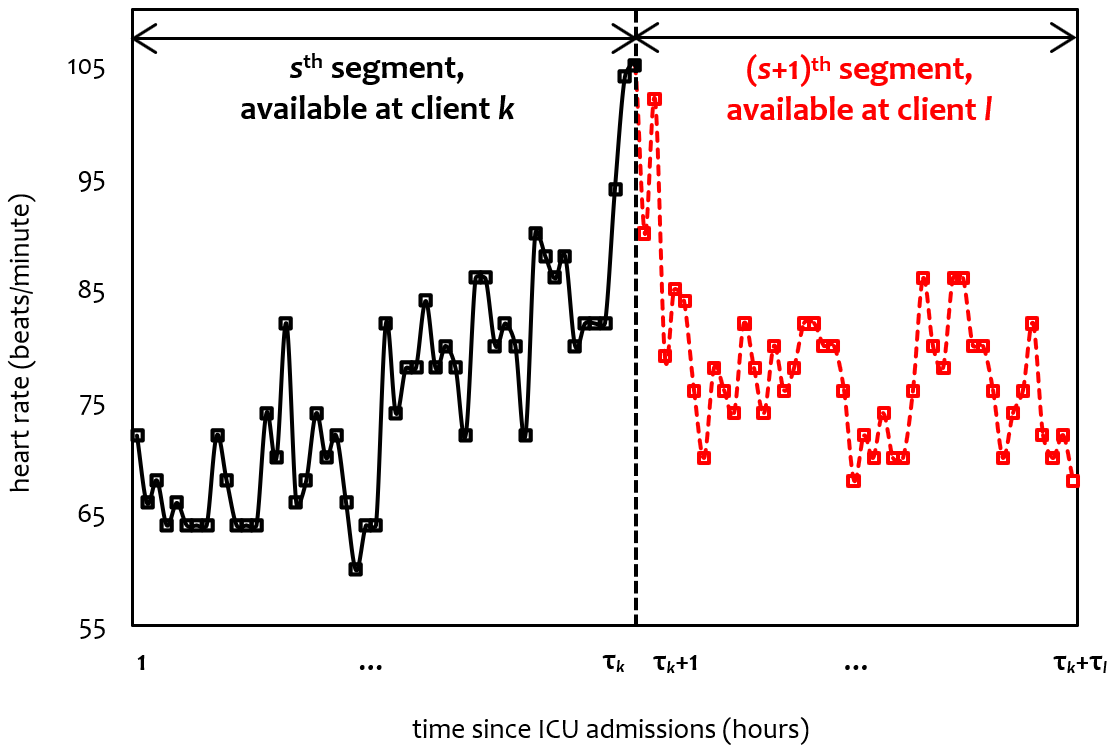}
  \caption{An example of a sequentially partitioned data distributed among clients, (artificially generated) heart rate of one patient (one training sample \(X_j\)) admitted to two different hospitals (two clients, \(k\) and \(l\)). The $s^{\textrm{th}}$ segment (time steps 1 to \(\tau_k\)) of \(X_j\) is generated and is available in client \(k\), and the ${(s+1)}^{\textrm{th}}$ segment (time steps \(\tau_k+1\) to \(\tau_k+\tau_l\)) of \(X_j\) is in client \(l\). see Section \ref{sec:pd}}
  \label{fig:fig2}
\end{figure}

\subsection{Problem Definition} \label{sec:pd}
As an example of a system generating privacy-sensitive sequentially partitioned data, suppose hospitals as clients participating in collaborative ML. Multiple-segment sequences of vital signs of each single patient is considered as one sequential training sample with a unique patient/sample ID. Consecutive segments of a sequential training sample are distributed among hospitals/clients, if the corresponding patient has multiple admissions to the hospitals. Figure \ref{fig:fig2} exemplifies a sequentially partitioned data, heart rate of one patient (one training sample, $X_j$) admitted to two different hospitals (two clients). The $s^{\textrm{th}}$ segment (time steps 1 to $\tau_k$) of the sequential training sample $X_j$ is generated and is available in client $k$, and the second segment (time steps $\tau_k+1$ to $\tau_k+\tau_l$) of $X_j$ is in client $l$.

In general, there are \(K\) clients generating consecutive segments of multiple-segment sequential data. The $j^{\textrm{th}}$ distributed multiple-segment sequential data which its two consecutive segments \(k\) and \(l\) are available at clients \(k\) and \(l\), is denoted by \(X_j=\{...,X_{j_s}^k, X_{j_{s+1}}^l, ...\}\). The two consecutive segments in \(X_j\) are sequences of length \(\tau_k\) and \(\tau_l\), \(X_{j_s}^k=\{x_{j}^1, ..., x_{j}^{\tau_k}\}\), and \(X_{j_{s+1}}^l=\{x_{j}^{\tau_k+1}, ..., x_{j}^{\tau_k+\tau_l}\}\). \(X_j\) can be rewritten as

\begin{equation} \label{eq:2}
X_j=\{..., x_{j}^{1}, ..., x_{j}^{\tau_k}, x_{j}^{\tau_k+1}, ..., x_{j}^{\tau_k+\tau_l}, ...\}.
\end{equation}

The index \(j\) is the sample ID of the training sample \(X_j\). The index \(k\) is the segment ID of the segment \(X_{j_s}^k\) which means the $s^{\textrm{th}}$ segment is generated and is available at client \(k\), Figure \ref{fig:fig2}.

Each multiple-segment sequence \(X_j\) has a single label \(Y_j\), being available at the client \(l\), containing the last segment of the sequence. For instance, in the problem of in-hospital mortality prediction (see Section \ref{sec:eicu}), the last hospital knows if the patient is alive or dead.

An ID bank is available on the server, containing a set of sample IDs \(S_{\textrm{ID}}=\{0, 1, ..., j, ...\}\) and sets of segment IDs \(S_{\textrm{segment}_{j}}=\{0, 1, ..., k, ...\}\). These sets of IDs are updated at each round of communication between clients and the server when the new generated training samples on clients are used in the FL process, see Section \ref{sec:fedslalgorithm}. For instance, in the current round of FL, the segment \(X_{j_{s+1}}^l\) is generated in the client \(l\). The client \(l\) checks the sample ID \(j\) with the set of sample IDs \(S_{\textrm{ID}}\) in the ID bank:

If \(j\) is not in \(S_{\textrm{ID}}\), \(j\) is added to \(S_{\textrm{ID}}\), and \(X_{j_{s+1}}^l\) is considered as the first segment of the multiple-segment sequence \(X_j\). In addition, the set \(S_{\textrm{segment}_{j}}\) is created with one element \(l\), \(S_{\textrm{segment}_{j}}=\{l\}\), which means the training sample with ID \(j\) has had one segment so far, on client \(l\).

If \(j\) is in \(S_{\textrm{ID}}\), \(l\) is added to \(S_{\textrm{segment}_{j}}\), and \(X_{j_{s+1}}^l\) is considered as the latest segment (up to now) of the multiple-segment sequence \(X_j\). It should be noted that in the problem defined in this subsection, only sample IDs (\(j\)) and segment IDs (\(k\)) are shared between clients and the ID bank on the server, and no label sharing (\(Y_j\)) or raw-data sharing (\(X_j\)) is required.

\begin{figure}
\centering
  \includegraphics[scale=.36]{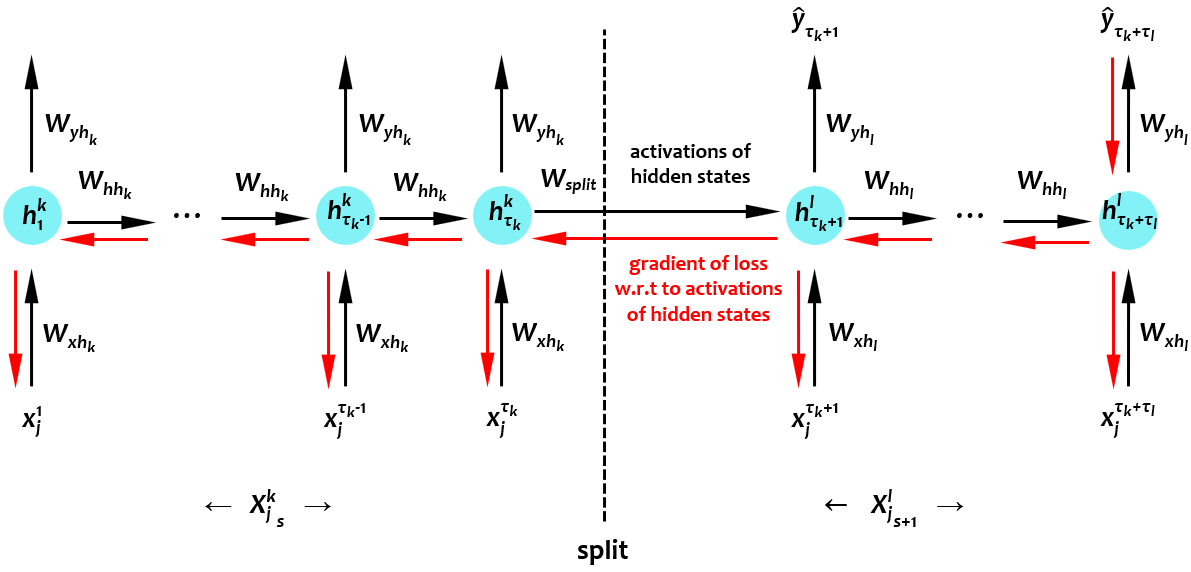}
  \caption{SL in RNNs. Two consecutive segments of a multiple-segment sequence \(X_j\) (\(X_{j_s}^k\) and \(X_{j_{s+1}}^l\)) are available at clients \(k\) and \(l\). From the dashed line, the RNN is split into two sub-networks. The left-side sub-network is trained on client \(k\), while the right-side sub-network is trained on client \(l\). The two sub-networks communicate to perform forward (black arrows) and backward (red arrows) propagations and update their weights. The left-side sub-network sends the activations of its hidden states to the right-side sub-network, while the right-side sub-network sends the gradient of the loss w.r.t. the activations of the hidden states of the left-side sub-network to the left-side sub-network. $x$'s, $y$'s, $h$'s, and $w$'s are inputs, outputs, hidden states, and weights of RNN at different time steps, respectively (see Section \ref{sec:slrnn}).}
  \label{fig:fig3}
\end{figure}

\subsection{Split Learning in Recurrent Neural Networks} \label{sec:slrnn}
Our aim is to train ML models, in a privacy-preserving setting, on multiple-segment sequential data with segments distributed among clients (see Section \ref{sec:pd}). The common ML models for dealing with sequential data are RNNs, such as the vanilla RNNs, LSTMs, GRUs, etc. \cite{sherstinsky2020fundamentals}. We introduce SL for RNNs. A RNN is split into sub-networks each being trained on one client and with one segment of multiple-segment training sequences.

In the previous SL algorithms (see Sections \ref{sec:sl} to \ref{sec:iflsl}), the split layer is a feed-forward layer. From a feed-forward layer, a feed-forward neural network is split between clients (containing training data) and server (containing training labels). However, RNNs have sequential (recurrent) structure to analyze sequential data. In the following, we explain how RNNs are split from recurrent layers.

In RNNs, recurrent connections between hidden states create a directed graph along a temporal sequence which enables RNNs to model temporal sequential behavior (see Figure \ref{fig:fig3}). Likewise, in our problem, segments of a multiple-segment sequential data (\(X_{j_s}^k\) and \(X_{j_{s+1}}^l\)) are temporally related and continuation of each other, all together constituting one sequence \(X_j\) (see Section \ref{sec:pd}). In order to model this temporal relation between segments, all segments should be given to one RNN. However, segments are available on different clients. In addition, in privacy-preserving SL algorithms, sharing segments between clients is not allowed. Therefore, A RNN (from the connections between hidden states) is split into sub-networks, each being trained on one client with one segment. The sub-networks communicate with each other through their hidden states in order to preserve the temporal relation between segments available on different clients.

Figure \ref{fig:fig3} illustrates how a multiple-segment sequential data, available on different clients (See Figure \ref{fig:fig2} and Equation \ref{eq:2}), is given to an (unsplit) RNN. \(X_j\) contains multiple segments, and its two consecutive segments \(X_{j_s}^k\) of length \(\tau_k\), and \(X_{j_{s+1}}^l\) of length \(\tau_l\) are available at clients \(k\), and \(l\), respectively. These two segments can be considered as a result of a split in \(X_j\) between time steps \(\tau_k\) and \(\tau_k+1\). Likewise, we split RNN from the hidden layer between hidden states \(h_{\tau_{k}}^k\) and \(h_{\tau_{k}+1}^l\). The split layer is indicated by a dashed line and its weight is named \(W_{\textrm{split}}\) in Figure \ref{fig:fig3}. This split results in two sub-networks, the sub-network at the left side of the split line is trained on client \(k\) with  \(X_{j_{s}}^k\), while the sub-network at the right side of the split line is trained on client \(l\) with \(X_{j_{s+1}}^l\).

The values of the hidden state, output, loss, and the gradient of loss w.r.t. \(W_{\textrm{split}}\) at time step \(\tau_{k}+1\), in the right-side sub-network (on client \(l\)) are computed as follows

\begin{equation} \label{eq:3}
h_{\tau_{k}+1}^l=\varphi_h(W_{\textrm{split}}\,h_{\tau_{k}}^k+W_{xh_l}\,x_j^{\tau_{k}+1})
\end{equation}

\begin{equation} \label{eq:4}
\hat{y}_{\tau_{k}+1}=\varphi_y(W_{hy_l}\,h_{\tau_{k}+1}^l)
\end{equation}

\begin{equation} \label{eq:5}
L_{\tau_{k}+1}=loss(\hat{y}_{\tau_{k}+1},\,{y}_{\tau_{k}+1})
\end{equation}

\begin{equation} \label{eq:6}
\frac{\partial L_{\tau_{k}+1}}{\partial W_{\textrm{split}}}=\frac{\partial L_{\tau_{k}+1}}{\partial \hat{y}_{\tau_{k}+1}}\,\frac{\partial \hat{y}_{\tau_{k}+1}}{\partial h_{\tau_{k}+1}^l}\,\frac{\partial h_{\tau_{k}+1}^l}{\partial W_{\textrm{split}}}
\end{equation}
\\
The first two derivatives in the right-hand side of Equation \ref{eq:6} are computable by the inputs, activations, weights, and the labels in the right-side sub-network (on client \(l\)). However, the third derivative \(\partial h_{\tau_{k}+1}^l\)/\(\partial W_{\textrm{split}}\) equals to \(h_{\tau_{k}}^k\,{{\varphi}^\prime}_h(W_{\textrm{split}}\,h_{\tau_{k}}^k+W_{xh_l}\,x_j^{\tau_{k}+1})\) and is dependent to \(h_{\tau_{k}}^k\) in the left-side sub-network. Therefore, the right-side sub-network needs the activation of hidden state of the left-side sub-network, \(h_{\tau_{k}}^k\). By the same token, the left-side sub-network needs the gradient of the loss in the right-side sub-network w.r.t. the activation of hidden state of the left-side sub-network, \(\partial L_{\tau_{k}+1}\)/\(\partial h_{\tau_{k}}^k\). Therefore, the two sub-networks need to communicate with each other to be able to perform forward and backward propagations, calculate the gradients, and update their parameters.

Note that the above gradient computations and the Equations \ref{eq:3} to \ref{eq:6} in split RNN are different from the computations in a conventional RNN resulting in the Back Propagation Through Time (BPTT) equations \cite{lillicrap2019backpropagation}. Nevertheless, the BPTT equations are true for each sub-network separately.

Based on the above computations, Algorithm 1 describes SL in RNNs. The input to the algorithm is multiple-segment sequential data \(X_j=\{...,X_{j_s}^k, X_{j_{s+1}}^l, ...\}\) with consecutive segments distributed among clients, and the labels \(Y_j\) being available at clients containing the last segment. Client \(k\) contains \(s^\textrm{th}\) segments \(X_{j_{s}}^k\) and trains sub-network \(W_s^k\), while client \(l\) contains \({(s+1)}^\textrm{th}\) segments \(X_{j_{s+1}}^l\) and trains sub-network \(W_{s+1}^l\). \(f_k\), \(f_k^\prime\), \(f_l\), and \(f_l^\prime\) denote forward and backward propagation computations in clients \(k\) and \(l\), respectively.

Without loss of generality, in Algorithm 1, we assume that \(X_j=\{X_{j_s}^k,\,X_{j_{s+1}}^l\}\), i.e., client \(k\) contains the starting segment of \(X_j\) with no previous segments, and client \(l\) contains the last segment of \(X_j\) with no next segments. While the vanilla RNN is considered in Algorithm 1, the proposed SL method is applicable to other RNNs, including GRUs, LSTMs, etc. Likewise, it is expandable to multi-layer and bidirectional RNNs. In this way, we will be able to train RNNs on sequentially partitioned data across clients without compromising privacy. Instead of sharing raw data or model parameters between clients, the only information being transmitted among clients is the activations of the hidden states and the gradient of the loss w.r.t. the activations of the hidden states.

\subsubsection{Byproducts of SL in Recurrent Neural Networks} \label{sec:bp}
In addition to preserving the privacy of raw data, splitting RNNs have the following benefits.


(i) Large number of time steps in a sequence is challenging to model for RNNs \cite{luo2020dualpath}. By splitting RNNs between clients, split RNNs deal with short segments of sequences, instead of long complete sequences

(ii) Proper hidden-state initialization is crucial in RNNs for successful sequence modeling \cite{mohajerin2019multistep, le2015simple}. In the proposed SL approach, the hidden states of the RNNs train on the next segments are initialized with the hidden states of the RNNs train on previous segments. This initialization strategy works better than random initialization since both previous and next RNNs are trained by the consecutive time steps of the same sequence.

\subsection{Federated Split Learning on Distributed Sequential Data} \label{sec:fedslalgorithm}
In the previous subsections, we defined the problem of sequentially partitioned data among clients, and presented SL in RNNs. In this subsection, we present a distributed privacy-preserving ML framework utilizing FL along with SL in RNNs, FedSL, to train models on sequentially partitioned data.

\begin{algorithm}[H]
 \textbf{While} client \(k\) has new \(s^\textrm{th}\) segments and client \(l\) has new \({(s+1)}^\textrm{th}\) segments of sequential training data
 \textbf{do}
  \begin{quoting}
  \textbf{Client \(\textbf{\emph{k}}\) executes:}
  \\\quad1- Initialize weights \(W_s^k\) randomly
  \\\quad2- Initialize hidden state \(h_{\tau_k}^k\) randomly
  \\\quad3- Compute forward propagation to get the new hidden state value, \(h_{\tau_k}^k\leftarrow\)\(f_k(X_{j_{s}}^k,\,h_{\tau_k}^k,\,W_s^k)\)
  \\\quad4- Send \(h_{\tau_k}^k\) to client \(l\)
\\
  \textbf{Client \(\textbf{\emph{l}}\) executes:}
  \\\quad5- Initialize weights \(W_{s+1}^l\) randomly
  \\\quad6- Initialize hidden state \(h_{\tau_{k+1}}^l\) with the received hidden state from client \(k\),  \(h_{\tau_{k+1}}^l\)\(\leftarrow\)\(h_{\tau_{k}}^k\)
  \\\quad7- Compute forward propagation to get the output at last time step, \(\hat{y}_{\tau_{k}+\tau_{l}}\)\(\leftarrow\)\(f_l(h_{\tau_{k+1}}^l,\,X_{j_{s+1}}^l,\,W_{s+1}^l)\)
  \\\quad8- Having the label, compute the value of loss at the last time step, \(L_{\tau_{k}+\tau_{l}}=loss(\hat{y}_{\tau_{k}+\tau_{l}},\,{y}_{\tau_{k}+\tau_{l}})\)
  \\\quad9- Compute the gradient of loss w.r.t \(W_{s+1}^l\), \(\frac{\partial L_{\tau_{k}+\tau_{l}}}{\partial W_{s+1}^l}\)\(\leftarrow\)\(f_l^{\prime}(h_{\tau_{k+1}}^l,\,X_{j_{s+1}}^l,\,W_{s+1}^l)\)
  \\\quad10- Update \({(s+1)}^\textrm{th}\) model parameters in client \(l\), \(W_{s+1}^l\)\(\leftarrow\)\(W_{s+1}^l-\eta\frac{\partial L_{\tau_{k}+\tau_{l}}}{\partial W_{s+1}^l}\) (\(\eta\) is the learning rate)
  \\\quad11- Compute the gradient of loss w.r.t the hidden state in client \(k\), \(\frac{\partial L_{\tau_{k}+\tau_{l}}}{\partial h_{\tau_k}^k}\)
  \\\quad12- Send \(\frac{\partial L_{\tau_{k}+\tau_{l}}}{\partial h_{\tau_k}^k}\) to client \(k\)
  \\\textbf{Client \(\textbf{\emph{k}}\) executes:}
  \\\quad13- Having \(\frac{\partial L_{\tau_{k}+\tau_{l}}}{\partial h_{\tau_k}^k}\), compute the gradient of loss at last time step on client \(l\) w.r.t \(W_{s}^k\), \(\frac{\partial L_{\tau_{k}+\tau_{l}}}{\partial W_{s}^k}\)\(\leftarrow\)\(f_k^{\prime}(X_{j_{s}}^k,\,h_{\tau_k}^k,\,W_{s}^k)\)
  \\\quad14- Update \(s^\textrm{th}\) model parameters in client \(k\), \(W_{s}^k\)\(\leftarrow\)\(W_s^k-\eta\frac{\partial L_{\tau_{k}+\tau_{l}}}{\partial W_s^k}\) (\(\eta\) is the learning rate)
  \end{quoting}
 \caption{Split Learning for Recurrent Neural Networks}
\end{algorithm}

\begin{figure}
\centering
  \includegraphics[scale=.18]{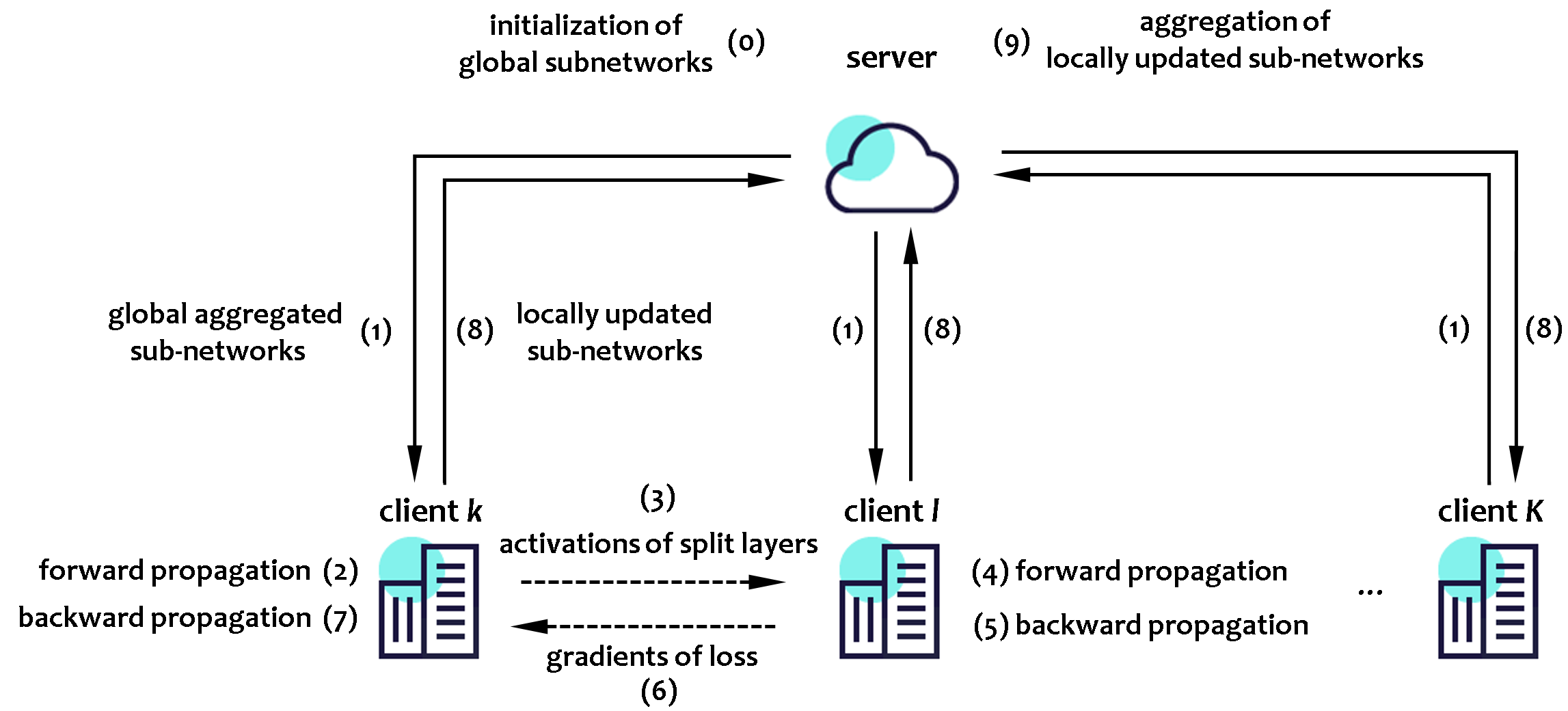}
  \caption{Block diagram of the proposed federated split learning algorithm, FedSL, to train models on sequentially partitioned data. See Section \ref{sec:fedslalgorithm}. The numbers (0)-(9) are correspondent to the numbers in Algorithm 2.}
  \label{fig:fig4}
\end{figure}

There are \(K\) clients in the federated split learning environment. At each round \(t\), \(t = 0,\) …,\( T\), \(C_t\) fraction of clients are selected by the server to participate in federated split learning. \(m_t = \textrm{max} (C_t . K,\,1)\) is the number of clients participating in the round \(t\) of algorithm. \(bs\), and \(ep\) are local mini-batch size and the number of local epochs on clients for local training. \(S\) is the maximum number of segments in multiple-segment sequential data.

\({W_{s}^k}^{(t)}\), \(s = 1,\) …, \(S\), \(k = 1,\) …, \(K\), and \(t = 0,\) …, \(T\) is the model trained locally at client \(k\) with \(s^\textrm{th}\) segments of sequential data at  round \(t\) of FedSL. \(W_s^{(t)}\) is the model generated on the server by aggregating locally trained models \({W_{s}^k}^{(t)}\) at the \(t^\textrm{th}\) round of communication between clients and the server.

Detailed steps of the proposed method are given in Algorithm 2 as well as in Figure \ref{fig:fig4}. The steps of the algorithm in Figure \ref{fig:fig4} are correspondent to the steps in Algorithm 2.

\begin{algorithm}
\textbf{Input:} multiple-segment sequential data distributed between clients, \(\{X_j, Y_j\}\)\\
\textbf{Output:} trained split RNNs, \(W_s, s = 1,\) …, \(S\)\\

\textbf{(0)} Server initializes models \(W_s^{(t)}\), \(s = 1,\) …, \(S, t = 0\)\\
\textbf{for} each round \(t\) = 1, 2, ..., \(T\) \textbf{do}
\begin{quoting}
\textbf{(1)} Server sends \(W_s^{(t)}\), \(s = 1, \)…, \( S\) to all \(m_t\) clients participating in the round \(t\) of federated split learning\\
\textbf{(2)} Each client \(k\), containing the \(s^\textrm{th}\) segment of \(X_j\) (\(X_{j_s}^k\)), performs forward propagation on the model \(W_s^{(t)}\) up to its split layer\\
\textbf{(3)} Each client \(k\), performed forward propagation using \(s^\textrm{th}\) segment in (2), sends the activation of its split layer to client \(l\), containing the \({(s+1)}^\textrm{th}\) segment\\
\textbf{(4)} Each client \(l\), having \({(s+1)}^\textrm{th}\) segment of \(X_j\) (\(X_{j_{s+1}}^k\)), the label \(Y_j\), and received the activation from the client \(k\) in (3), performs forward propagation on \(W_{s+1}^{(t)}\) from its split layer to the final layer, and computes the output and the loss values\\
\textbf{(5)} Each client \(l\), performed forward propagation in (4), performs backward propagation on \(W_{s+1}^{(t)}\), and updates its weights to generate split model \({W_{s+1}^l}^{(t+1)}\)\\
\textbf{(6)} Each client \(l\), performed backward propagation in (5), sends the gradient of its loss value, w.r.t to the activation received from the client \(k\), to the client \(k\)\\
\textbf{(7)} Each client \(k\), received the gradient of the loss from the client \(l\) in (6), performs backward propagation on \(W_{s}^{(t)}\) , and updates its weights to generate split model \({W_{s}^k}^{(t+1)}\)\\
\textbf{(8)} All clients send their locally-updated split models to the server\\
\textbf{(9)} Sever aggregates all received split models to generate global models for each segment ID as follows\\
\(W_{s}^{(t+1)}=\sum _{k=1}^{K}\frac{n_{s}^k}{n_s}{W_{s}^k}^{(t+1)}\)\\
where \(n_{s}^k\) is the number of \(s^\textrm{th}\) segments of training samples at client \(k\), and \(n_s\) is the total number of \(s^\textrm{th}\) segments of training samples in all clients.
\end{quoting}

 \caption{Federated Split Learning on Distributed Sequential Data. The steps (2)-(7) are split learning between clients (see Algorithm 1)}
\end{algorithm}

\section{Experimental Results} \label{sec:er}
As per previous works in the area of FL and SL \cite{he2020fedml, gupta2018distributed, mcmahan2017communicationefficient, singh2019detailed, thapa2020splitfed, chen2020vafl, vepakomma2018split, yin2021comprehensive, hanaccelerating, mugunthan2021multi}, we evaluate the performance of FedSL on image classification and electronic health record datasets. We perform experiments using different variations of RNNs, in different settings, and in comparison with previous related works. We implemented the experiments in PyTorch on a server with 64 GB of RAM and NVIDIA Tesla P100 PCIe 12 GB GPU. The code of our implementations is available at https://github.com/abedicodes/FedSL.

\subsection{(Sequential) Image Classification}
The first dataset is sequential MNIST dataset \cite{le2015simple} which is constructed from the original MNIST handwritten digit recognition dataset by converting \(28\times28\) pixel gray-level images into 784 sequences, i.e. the models read one pixel at a time in scan-line order, starting at the top left corner of the image, and ending at the bottom right corner, \cite{le2015simple}. This dataset contains 60,000 training and 10,000 test samples. There is an equal number of samples in each of ten digit classes.

The second dataset is fashion MNIST dataset which has been used extensively in the literature for evaluating FL algorithms, \cite{he2020fedml, thapa2020splitfed}. This dataset contains equal number of images of ten different types of clothes. There are 60,000 training and 10,000 test gray-level images of size  \(28\times28\). The samples in this dataset are images. However, RNNs in the proposed method deal with distributed sequential data. Therefore, in our experiments, each \(28\times28\) pixel image is converted to a sequence of length 28 with 28-element feature vector in each time step, i.e. each 28-pixel row of the image is considered as one time step of a 28-length sequence.

The above datasets are not sequentially partitioned or distributed. As per the previous works in the area of FL and SL \cite{mcmahan2017communicationefficient, he2020fedml, gupta2018distributed}, we will divide training samples into segments and distribute the segments among consecutive clients in a federated setting. The performance metrics for image classification datasets are training loss and test-set accuracy in communication rounds of FL algorithms.

\subsubsection{(Sequential) MNIST}
The sequential MNIST dataset is used to evaluate the performance of the proposed method in modeling long-range dependencies in long sequences, \cite{le2015simple, bai2018empirical}. Le et al. \cite{le2015simple} proposed to use ReLU as activation function in the vanilla RNN and initialize the hidden states with the identity matrix, and called it IRNN. The IRNN achieved superior performance compared to the vanilla RNN and LSTM in learning long-term dependencies. For the same reason, we use IRNN for sequential MNIST dataset classification.

In FedAvg \cite{mcmahan2017communicationefficient} (described in Section \ref{sec:fl}), which cannot handle sequential data, complete training sequences are available at single clients, i.e. clients contain complete 784-length sequences. However, in the proposed FedSL, each 784-length sequence is distributed among two or three clients, i.e. every two or three consecutive clients contain consecutive segments of the complete sequences. In both the FedAvg and FedSL approaches, at each time step of sequences, a single-element feature vector is available corresponding to one pixel in the MNIST images.

In FedAvg, the IRNN has one unidirectional layer with 64 neurons, and after that a fully-connected layer with 64 neurons. The hidden states are initialized with the identity matrix. In FedSL, the IRNN is split into two or three sub-networks. The first sub-networks, trained on the clients containing the first segments of training sequences, are IRNNs having one unidirectional layer with 64 neurons. The hidden states are initialized with the identity matrix. The second sub-networks, trained on the clients containing the second segments of training sequences,  are IRNNs with one unidirectional layer with 64 neurons. The hidden states of the second sub-networks are initialized by the first sub-networks during communications in SL (see Section \ref{sec:slrnn}). The third sub-networks, trained on the clients containing the third segments of training sequences and the labels,  are IRNNs with one unidirectional layer with 64 neurons plus a fully-connected layer with 64 neurons. The hidden states of the third sub-networks are initialized by the second sub-networks during communications in SL (see Section \ref{sec:slrnn}). The learning rate in both methods is 0.000001.

\begin{figure}[ht]

\begin{subfigure}[b]{.5\textwidth}
\centering
  \includegraphics[scale=.14]{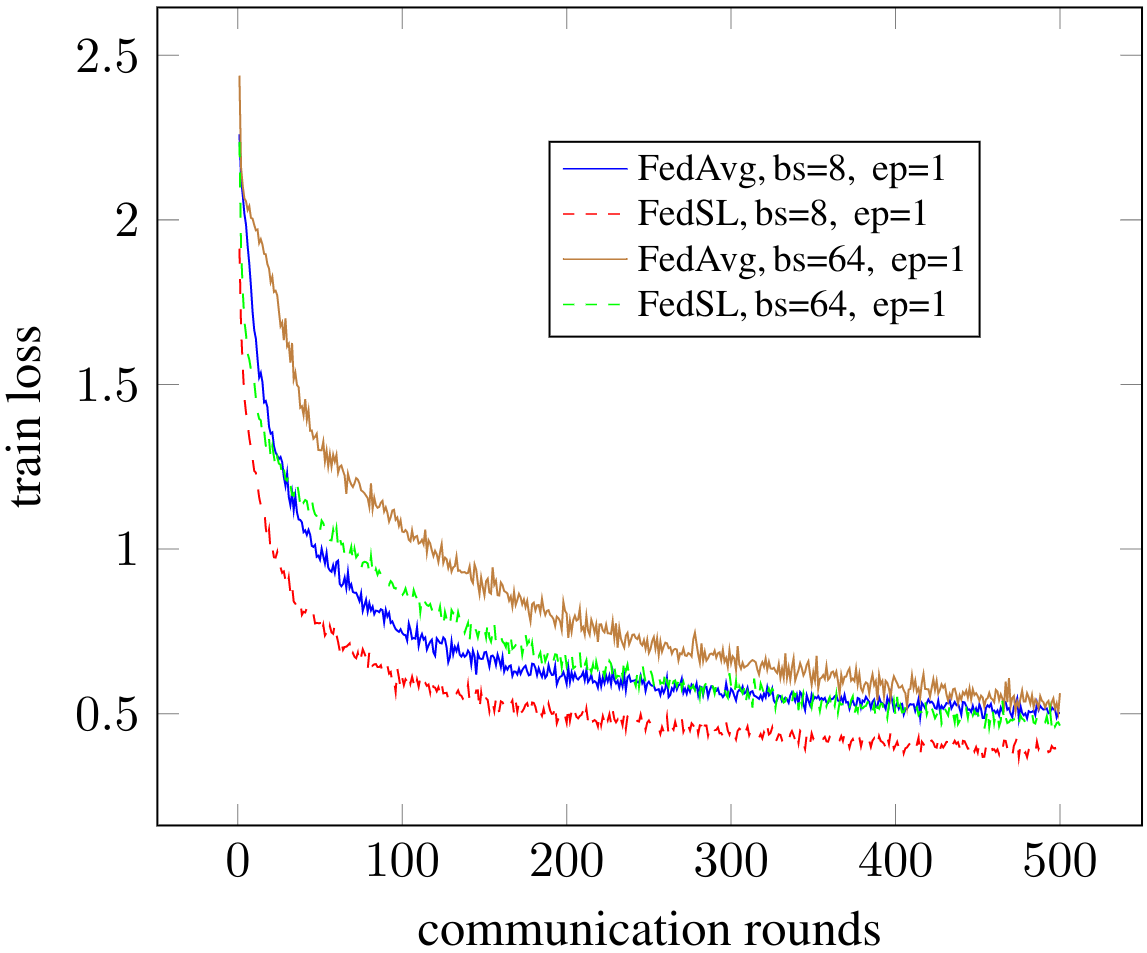}
\end{subfigure}
\begin{subfigure}[b]{.5\textwidth}
\centering
  \includegraphics[scale=.14]{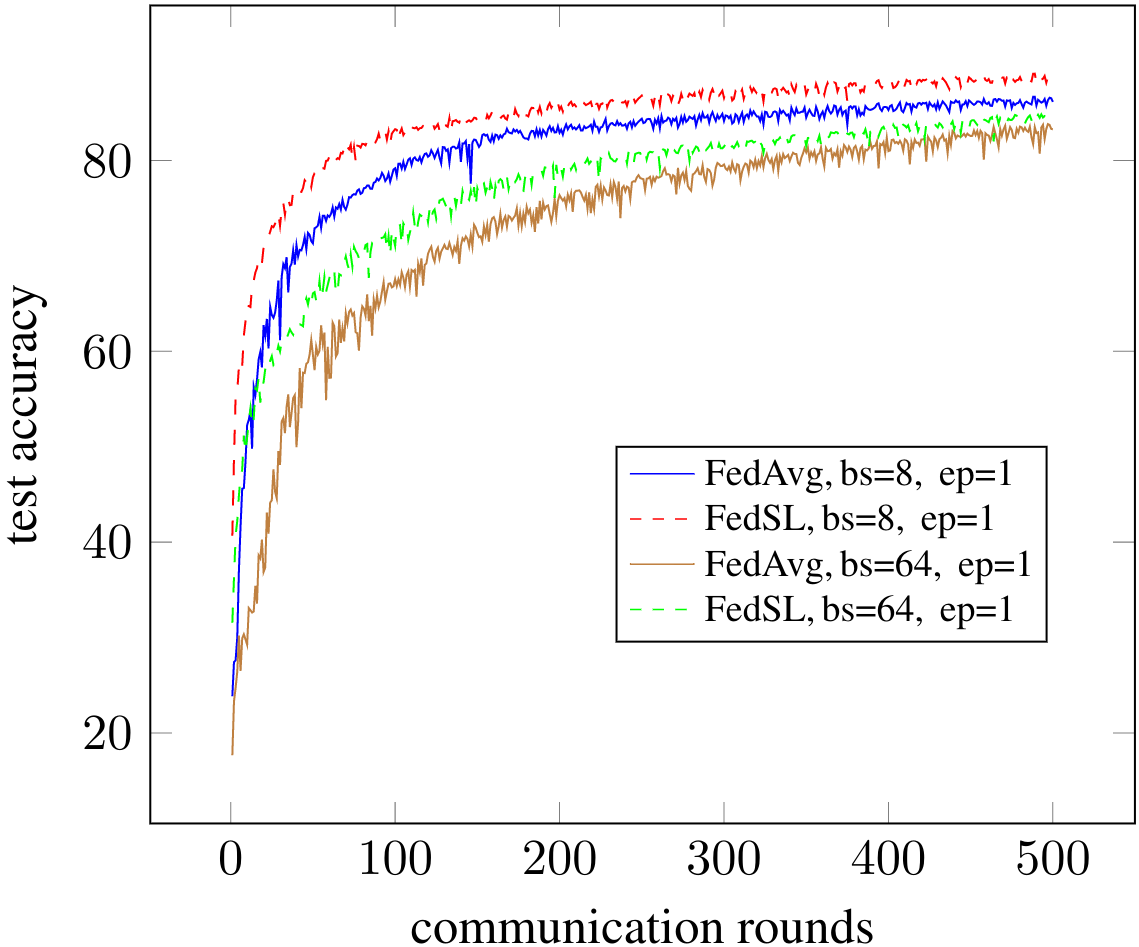}
\end{subfigure}

\caption{Train loss (left) and test accuracy (right) versus communication rounds for FedAvg \cite{mcmahan2017communicationefficient} and the proposed FedSL on the sequential MNIST dataset with different values of local batch sizes (bs). In all cases, the proposed FedSL achieves lower train loss and higher test accuracy in fewer rounds of communication.}
\label{fig:fig5}
\end{figure}

\noindent\textbf{Local computation per client:} Figure. \ref{fig:fig5} shows the learning curves of FedAvg and FedSL over 500 rounds of communication, when local batch size is 8, and 64. Train loss and test accuracy vs. the number of communication rounds are shown in Figure. \ref{fig:fig5} left and right, respectively. In this experiment, the distribution of data between clients is IID. The total number of clients \(K\) = 100 and \(C_t=0.1\), i.e. in each round of communication, \(m_t = \textrm{max} (0.1 \times 100,\,1)=10\) clients participate in federated averaging. As can be seen in Figure. \ref{fig:fig5}, in all cases, the proposed FedSL outperforms FedAvg, while handling distributed sequential data among clients. The FedSL achieves higher accuracy in fewer rounds of communication. This superiority of the proposed method is due to the byproducts of SL in RNNs in addition to preserving privacy (see Section \ref{sec:bp}). In both methods, increasing local computation per client (by decreasing the local batch sizes) results in convergence in fewer rounds of communication.

\begin{figure}[ht]

\begin{subfigure}[b]{.5\textwidth}
\centering
  \includegraphics[scale=.14]{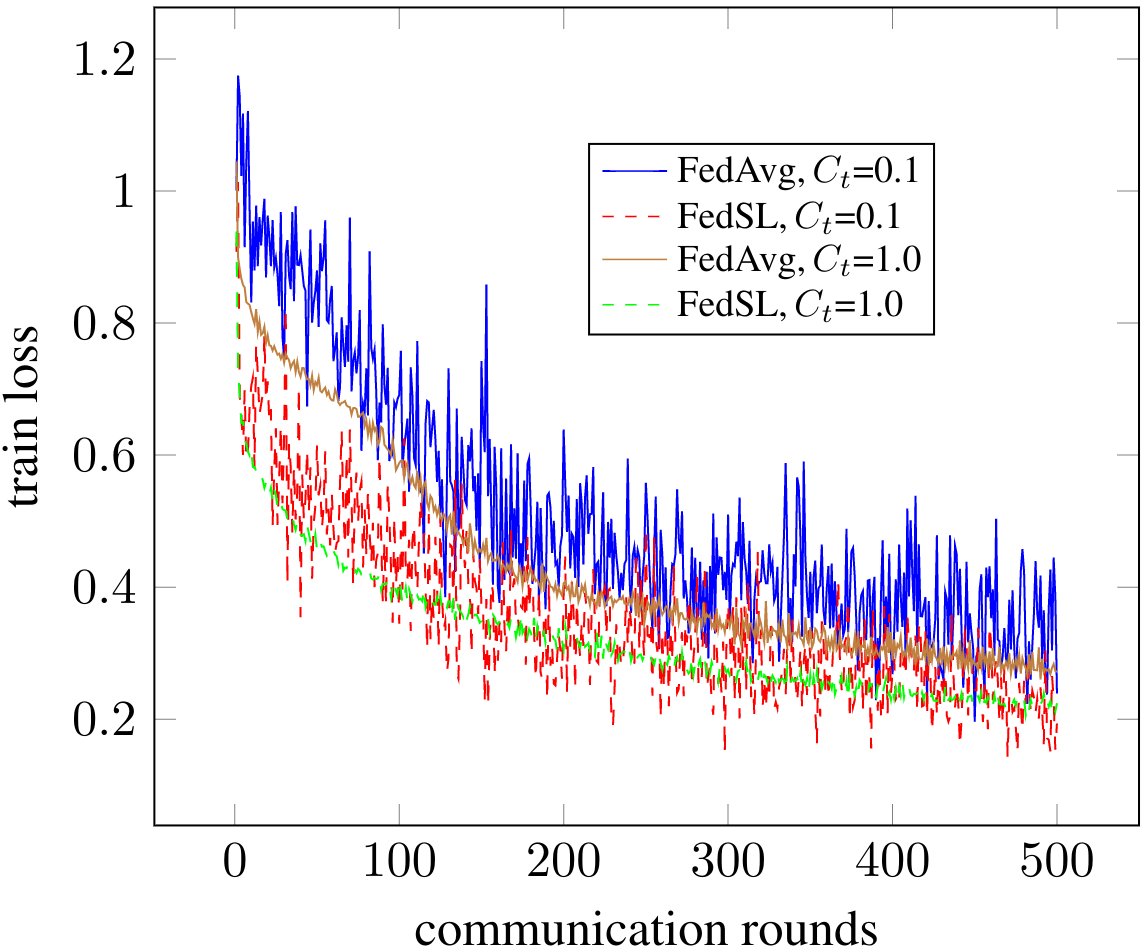}
\end{subfigure}
\begin{subfigure}[b]{.5\textwidth}
\centering
  \includegraphics[scale=.14]{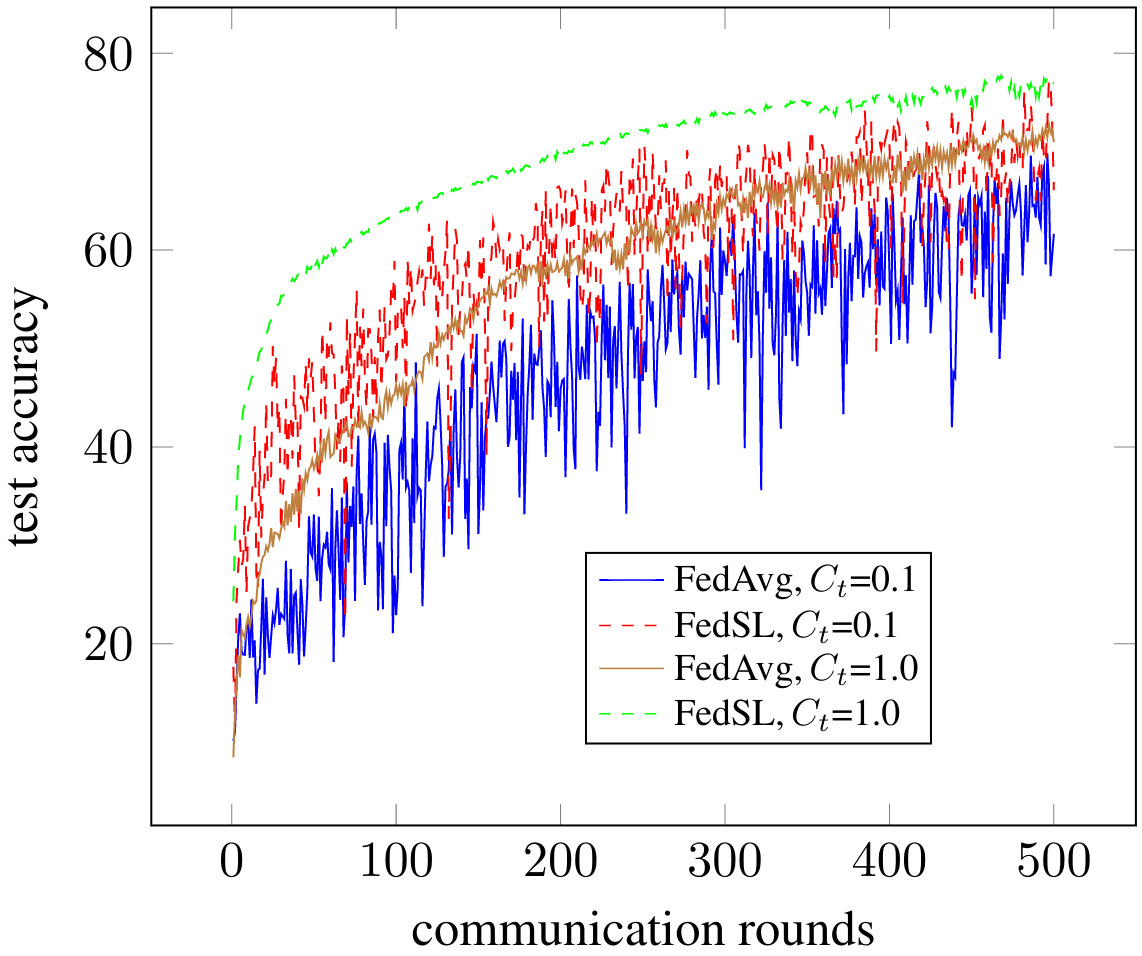}
\end{subfigure}

\caption{Train loss (left) and test accuracy (right) versus communication rounds for FedAvg \cite{mcmahan2017communicationefficient} and the proposed FedSL on the sequential MNIST dataset with non-IID distribution for different values of \(C_t\), different number of participants in federated learning process. In all cases, the proposed FedSL achieves lower train loss and higher test accuracy in fewer rounds of communication.}
\label{fig:fig6}
\end{figure}

\noindent\textbf{Increasing parallelism in non-IID data:} Figure. \ref{fig:fig6} compares FedSL and FedAvg in handling non-IID data, as described in \cite{mcmahan2017communicationefficient}, when different number of clients participate in FL, \(C_t\) is 0.1 and 1 (\(K\) = 100). In these experiments, local batch sizes, and local epochs equal 64, and 1, respectively. In all cases, FedSL achieves lower loss and higher accuracy in fewer rounds of communication compared to FedAvg, while handling multiple-segment sequential data. In both the methods, increasing parallelism, participating more clients in FL results in achieving lower train loss and higher test accuracy in fewer rounds of communications.

\begin{figure}[ht]

\begin{subfigure}[b]{.5\textwidth}
\centering
  \includegraphics[scale=.14]{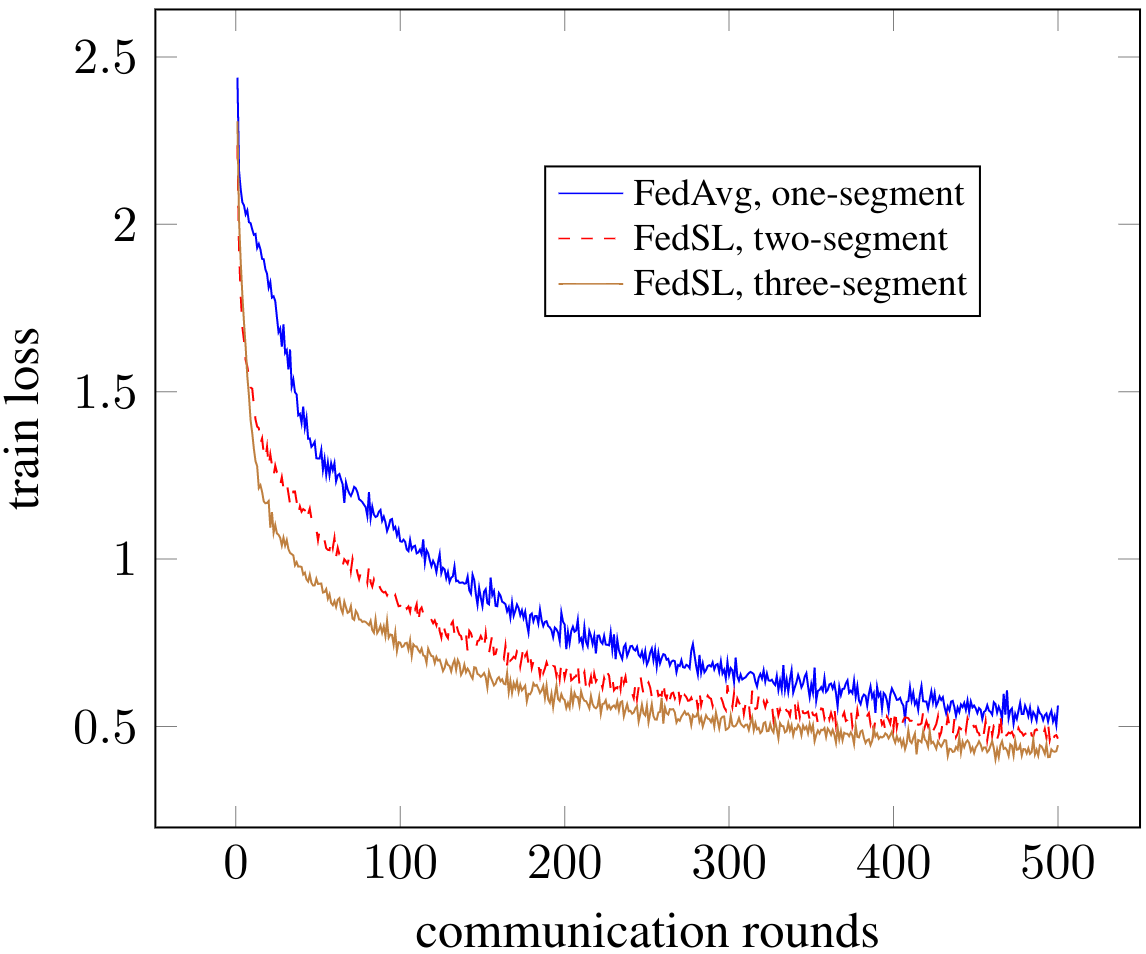}
\end{subfigure}
\begin{subfigure}[b]{.5\textwidth}
\centering
  \includegraphics[scale=.14]{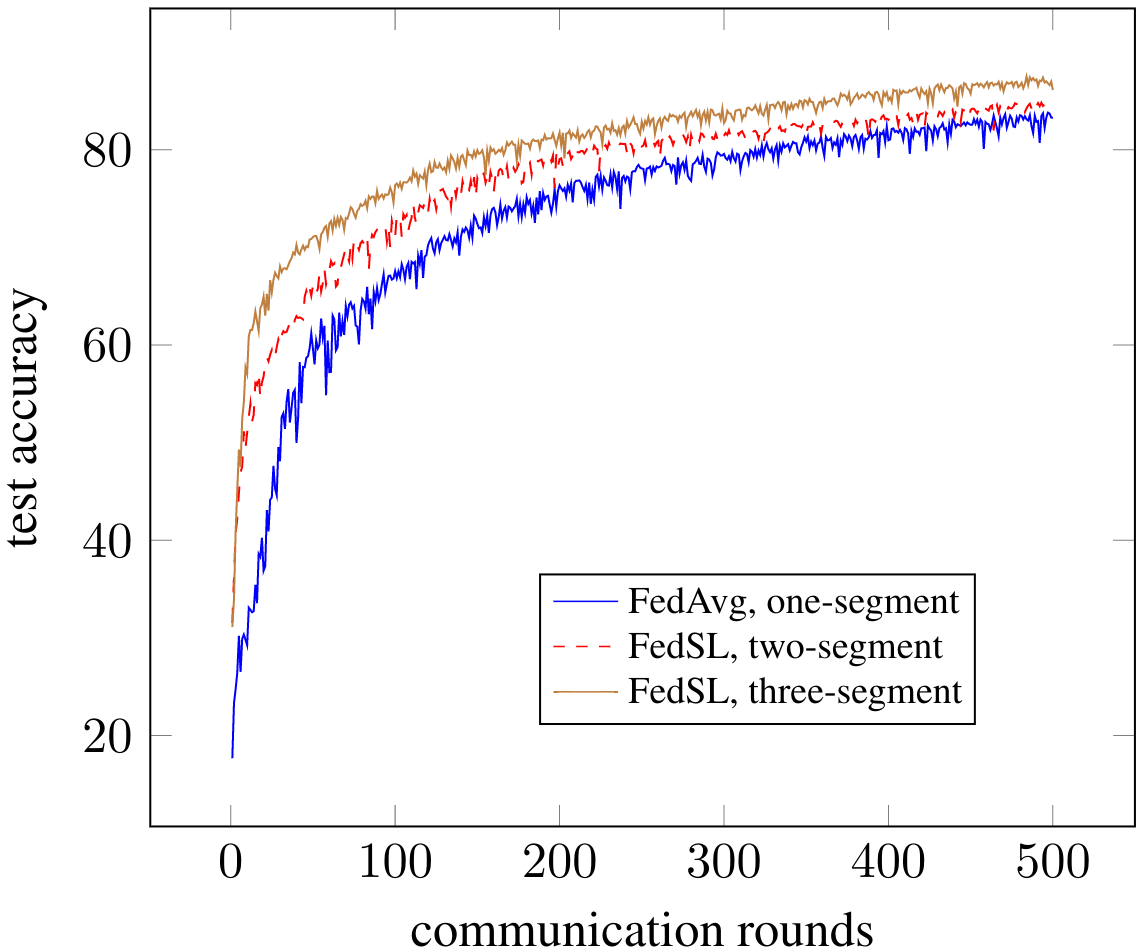}
\end{subfigure}

\caption{Train loss (left) and test accuracy (right) versus communication rounds for FedAvg \cite{mcmahan2017communicationefficient} (with one-segment undistributed data) and FedSL (with two and three-segment distributed data) using sequential MNIST dataset. There is an improvement in performance from one-segment sequential data (using FedAvg) to two and three-segment distributed sequential data (using FedSL).}
\label{fig:fig7}
\end{figure}

\noindent\textbf{Different number of distributed segments:} In the two previous experiments on FedSL, two-segment sequential data were distributed among clients. In this experiment, we study the effect of the number of distributed segments among clients using sequential MNIST dataset. In undistributed case, the entire 784 time steps of sequences are available on single clients. In two-segment distribution case, two  392-length sequences are distributed among two consecutive clients, while, in three-segment distribution case, one 264-length sequence and two 260-length sequences are distributed among three consecutive clients. Figure. \ref{fig:fig7} depicts the learning curves of the above three cases in 500 rounds of communication using local batches of size 64 and 1 local epochs. In the first case, the FedAvg is used, because the data is not sequentially distributed. In the two-segment and three-segment cases, the FedSL with two and three sub-networks are used to handle sequentially distributed sequences. As can be seen in Figure. \ref{fig:fig7}, FedSL outperforms FedAvg in terms of achieving lower loss and higher accuracy in fewer rounds of communication. In addition, when three-segment sequences are distributed among clients (and three sub-networks are trained in SL), higher accuracy is achieved in comparison with two-segment sequence distribution where two sub-networks are trained in SL. This improvement from one-segment (using FedAvg) to three-segment sequences is due to the byproducts of SL in sequential neural networks in addition to preserving privacy (see Section \ref{sec:bp}).

\begin{figure}[ht]

\begin{subfigure}[b]{.5\textwidth}
\centering
  \includegraphics[scale=.14]{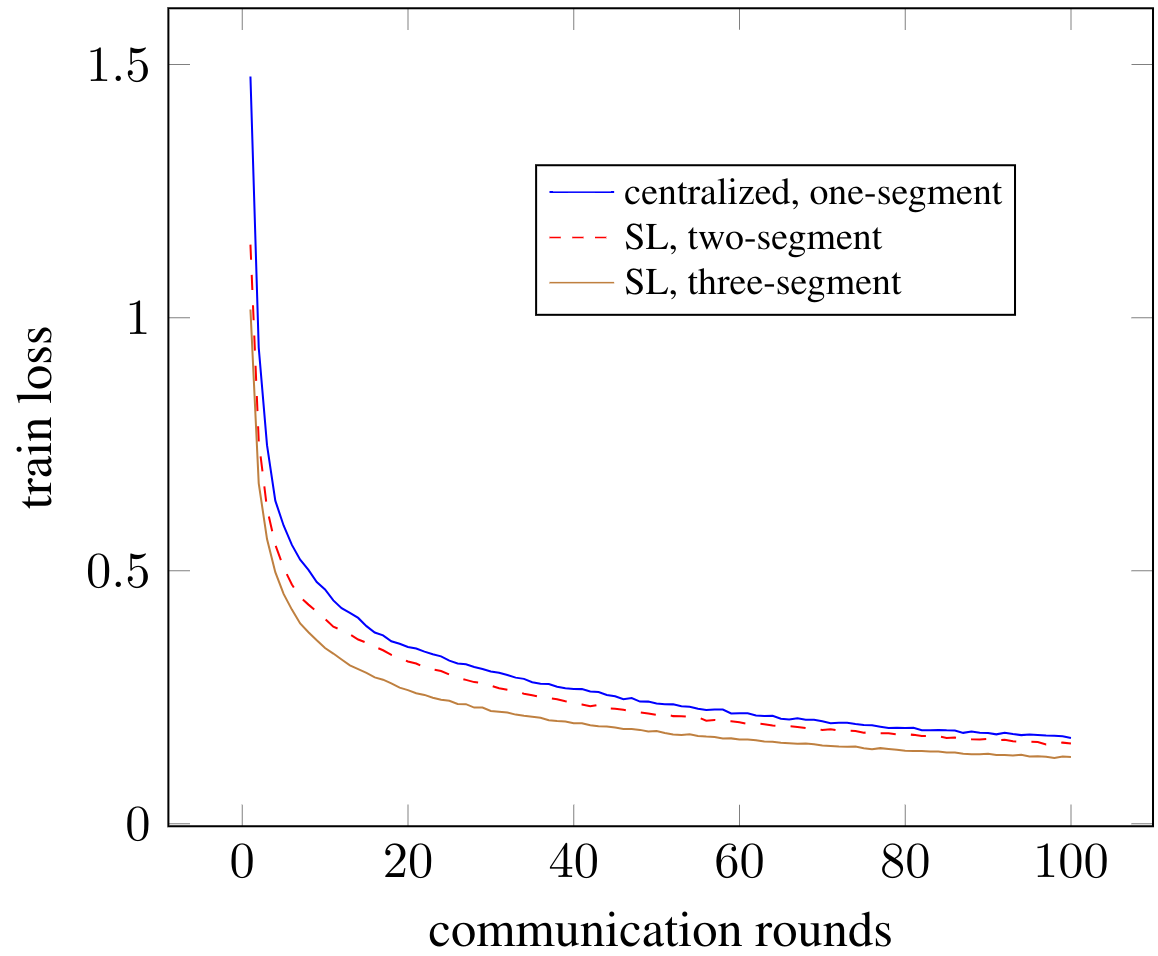}
\end{subfigure}
\begin{subfigure}[b]{.5\textwidth}
\centering
  \includegraphics[scale=.14]{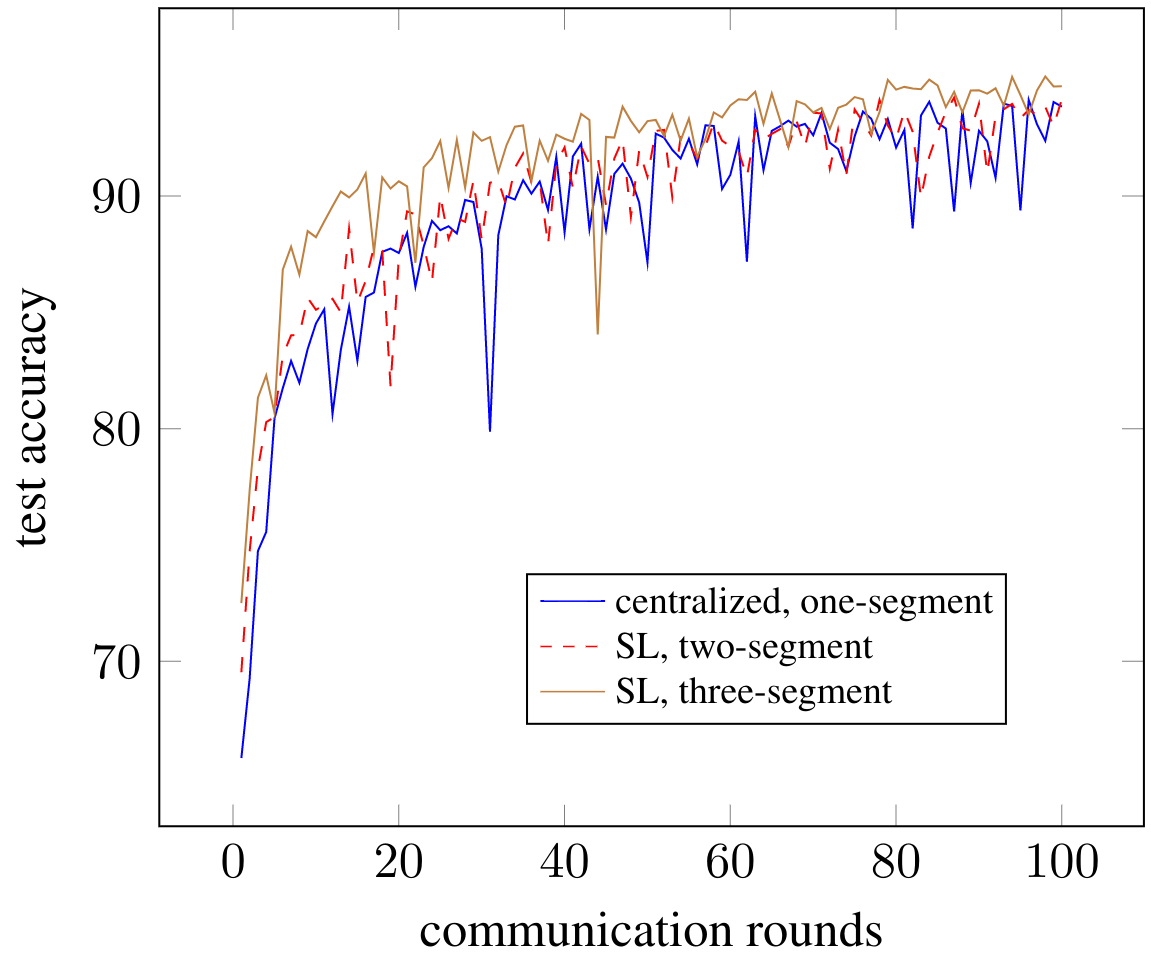}
\end{subfigure}

\caption{Train loss (left) and test accuracy (right) of applying centralized learning and the proposed SL approach (with two and three-segment distributed sequential data) to sequential MNIST dataset. Distributed learning using the proposed SL method for RNNs converges faster compared to centralized learning.}
\label{fig:fig8}

\end{figure}

\noindent\textbf{Comparing the proposed SL method for RNNs with centralized learning:} we examine the performance of the proposed SL algorithm for IRNNs, compared to the centralized learning. In SL, two (or three) consecutive clients contain one half (or one third) of the time steps of the training sequences. These two (or three) clients communicate to train their sub-networks without data sharing (see Section \ref{sec:slrnn}). Figure. \ref{fig:fig8} left and right depict the learning curves of the two methods when batch size in centralized learning and local batch sizes in clients in SL are 64. As can be observed in Figure. \ref{fig:fig8}, the proposed SL method for IRNNs outperforms the centralized learning with less convergence time.

\subsubsection{(Sequential) Fashion MNIST}
We report the results of applying the proposed method to the sequential fashion MNIST dataset, in comparison with FedAvg \cite{mcmahan2017communicationefficient} (described in Section \ref{sec:fl}). For the experiments related to the FedAvg, being unable to handle sequentially distributed data, it is assumed that complete training sequences are available at single clients, i.e. clients contain complete 28-length sequences. However, in the proposed FedSL, each 28-length sequence is distributed among consecutive two clients, i.e. each two consecutive clients contain 14-length segments of the complete sequences, one related to the first half and another is related to the second half of each 28-length sequence. In both FedAvg and FedSL approaches, at each time step of sequences, a 28-element feature vector is available corresponding to one row of the \(28\times28\) gray-level images. In this experiment, the GRU is used as sequence classifier. In FedAvg, the GRU has one unidirectional layer with 64 neurons, and after that a fully-connected layer with 64 neurons. The hidden states are initialized with zeros. In the FedSL, the GRU is split into two sub-networks. The first sub-networks, trained on the clients containing the first segments of training sequences, is a GRU having one unidirectional layer with 64 neurons whose hidden states are initialized with zeros. The second sub-networks, trained on the clients containing the second segments of training sequences and the label,  is a GRU with one unidirectional layer with 64 neurons plus a fully-connected layer with 64 neurons. The hidden states of the second sub-networks are initialized by the first sub-networks during communications in SL (see Section \ref{sec:sl}). The learning rate in both methods is 0.1.

\begin{figure}[ht]

\begin{subfigure}[b]{.5\textwidth}
\centering
  \includegraphics[scale=.14]{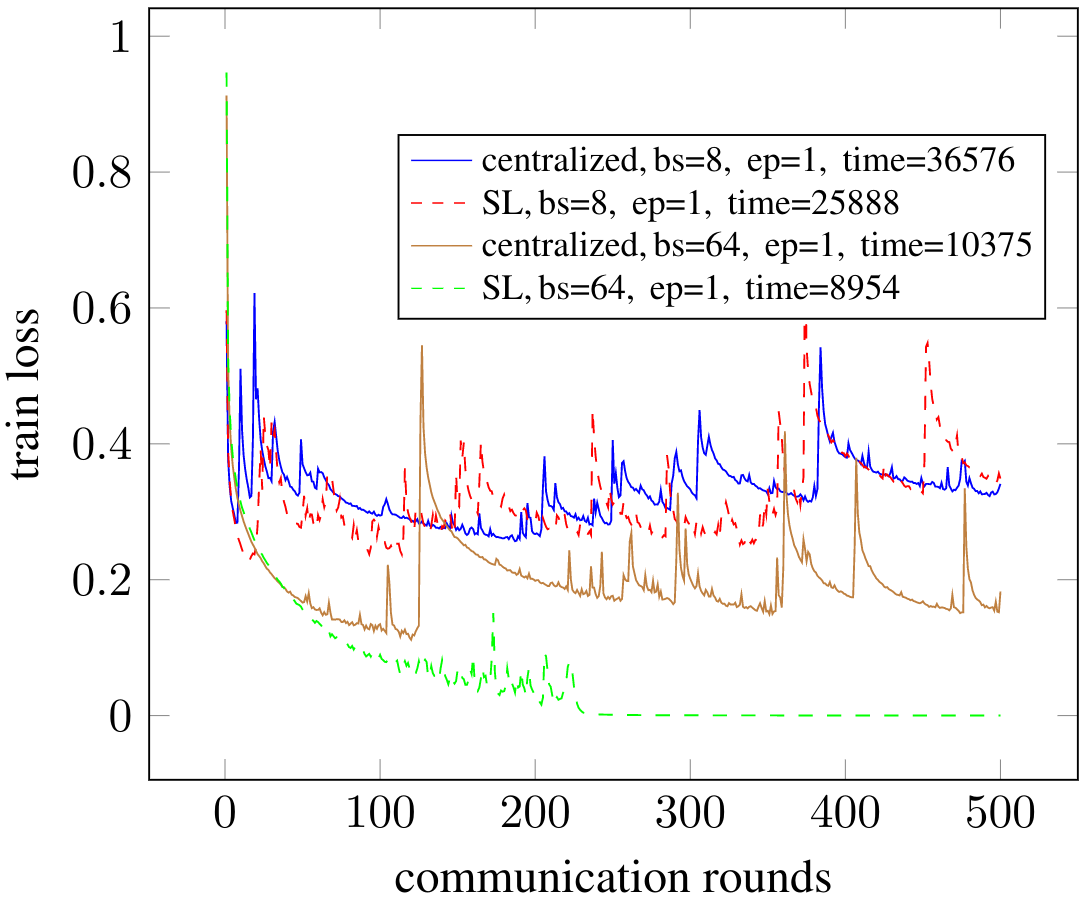}
\end{subfigure}
\begin{subfigure}[b]{.5\textwidth}
\centering
  \includegraphics[scale=.14]{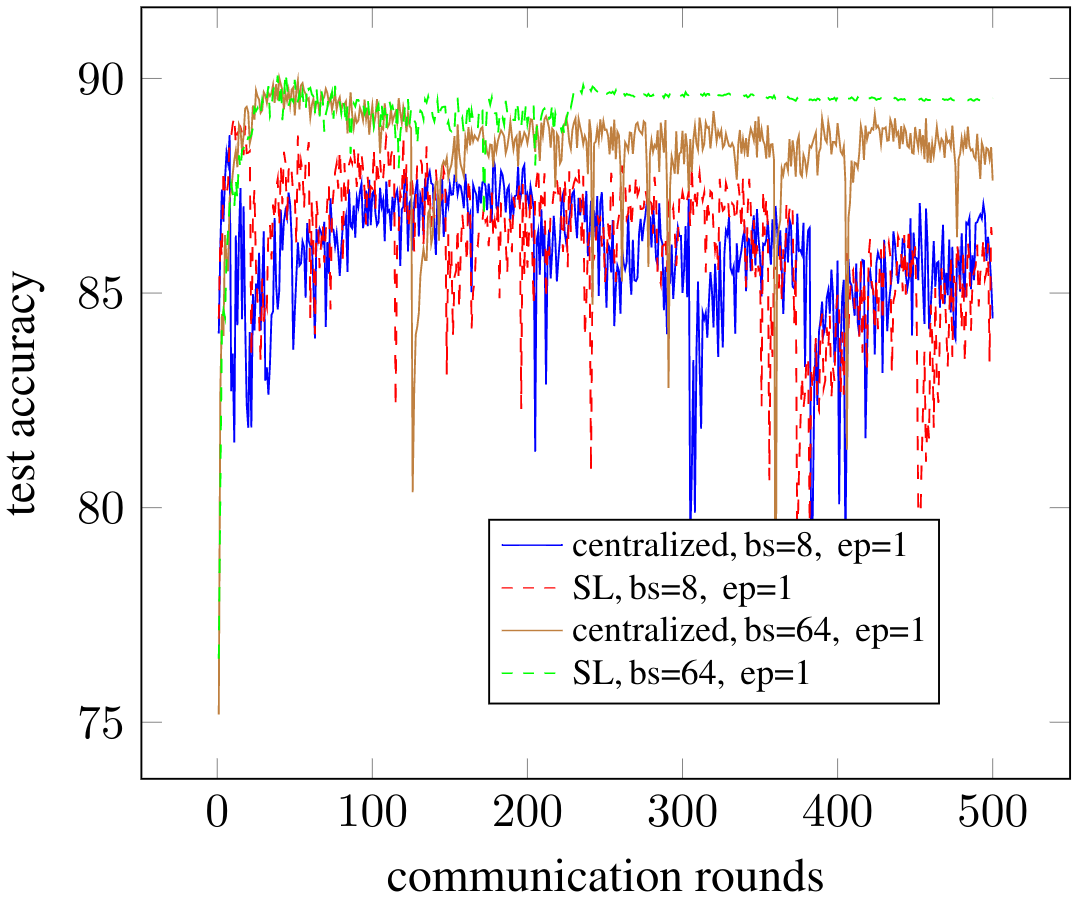}
\end{subfigure}

\caption{Train loss (left) and test accuracy (right) of applying centralized learning and the proposed SL approach to the sequential fashion MNIST dataset with different values of batch sizes. Distributed learning using the proposed SL method for RNNs converges faster compared to the centralized learning.}
\label{fig:fig11}
\end{figure}

\noindent\textbf{Comparing the proposed SL method for RNNs with centralized learning:} we examine the performance of the proposed SL algorithm for RNNs (GRU in this experiment), compared to the centralized learning. In the proposed SL environment, two consecutive clients contain one half of the training sequences. These two clients communicate to train their sub-networks without data sharing (see Section \ref{sec:sl}). Figure. \ref{fig:fig11} depicts learning curves of the two methods when batch size in centralized learning and local batch sizes in clients in SL are 8 and 64. The number of epochs in both methods is considered as 1. As can be observed in Figure. \ref{fig:fig11}, in all cases, the proposed SL method for RNNs outperforms centralized learning with much less convergence time and with smaller amount of fluctuations in the values of train loss.

\subsection{eICU Dataset} \label{sec:eicu}
The next dataset is eICU \cite{pollard2018eicu}, a multi-center Intensive Care Unit (ICU) public dataset containing data of 200,859 admissions to ICUs for 139,367 unique patients in 208 hospitals located throughout the United States. In this dataset, there are patients with multiple unit/hospital admissions. The dataset is deidentified, and includes vital sign measurements, care plan documentation, severity of illness measures, diagnosis information, treatment information, etc., \cite{pollard2018eicu}.

Out of 31 available tables in the eICU dataset \cite{pollard2018eicu}, we extract information from the following three tables, patient table (containing demographic and administrative information regarding the patient and their unit/hospital stay), lab table (containing laboratory measurements for patient derived specimens), and nurseCharting table (containing information charted at the bed side, including vital signs). As discussed in \cite{sheikhalishahi2020benchmarking}, 20 numerical and categorical variables are extracted as features describing the first 48 hours of ICU stays of patients. The 13 numerical features are scaled and transformed individually between -1 and 1, and the 7 categorical variables are one hot encoded, resulting in a 419-element feature vector for each hour of ICU stay. Therefore, we have a sequential data with 48 times steps with 419-element feature vector at each time step. In this paper, we work on in-hospital mortality prediction task which is a binary classification problem with labels 0 or 1 indicating survival or death of the patient after being discharged from the ICU. All patients with two admissions are extracted from all 208 hospitals resulting in 13,277 patients with two consecutive admissions with 11.57\% mortality rate. The label (survival or death of the patient) is available at the hospital containing the data of the second admission of the patient. Since, there are different values of mortality rate in different hospitals, this dataset is in nature non-IID \cite{huang2020loadaboost}. There are different numbers of samples in each of two classes in different clients. In our experiments, 80\% and 20\% of samples are used for training and test, respectively.

We report the results of applying the following methods to the eICU dataset: centralized learning, the proposed SL for RNNs (see Section \ref{sec:slrnn}), FedAvg \cite{mcmahan2017communicationefficient}, LoAdaBoost FedAvg \cite{huang2020loadaboost}, the proposed FedSL (see Section \ref{sec:fedslalgorithm}), and the combination of the proposed FedSL with LoAdaBoost FedAvg. In the centralized learning, all data is assumed to be available in one location and the RNN is trained using conventional centralized learning. In the proposed SL for RNNs (see Section \ref{sec:slrnn}), the segments of sequential data is assumed to be distributed among two consecutive clients. Two sub-networks are trained on two clients using consecutive segments of distributed sequential data. In FedAvg, all segments of each sequential training sample are assumed to be available in one client, full segment sequential data are distributed among clients. The LoAdaBoost FedAvg \cite{huang2020loadaboost}, described in Section \ref{sec:fl}, is an improved version of FedAvg that modifies the local updates at client side and can be incorporated with the proposed FedSL. In the combination of the FedSL with the LoAdaBoost FedAvg \cite{huang2020loadaboost}, the training of each sub-network in each client is performed through LoAdaBoost FedAvg \cite{huang2020loadaboost}.

For the eICU dataset, the LSTM is used as sequence classifier. In centralized learning, FedAvg, and the LoAdaBoost FedAvg \cite{huang2020loadaboost}, the LSTM has one unidirectional layer with 64 neurons, and after that a fully-connected layer with 64 neurons. The hidden states are initialized with zeros. In the proposed SL, FedSL, and the combination of FedSL with the LoAdaBoost FedAvg \cite{huang2020loadaboost}, the LSTM is split into two sub-networks. The first sub-networks, trained on the clients containing the first segments of training sequences, is an LSTM having one unidirectional layer with 64 neurons whose hidden states are initialized with zeros. The second sub-networks, trained on the clients containing the second segments of training sequences and the label, is an LSTM with one unidirectional layer with 64 neurons plus a fully-connected layer with 64 neurons. The hidden states of the second sub-networks are initialized by the first sub-networks during communications in SL (see Section \ref{sec:slrnn}). The learning rate for centralized learning and the proposed SL is 0.01, and for other methods is 0.1.

\begin{figure}[ht]

\begin{subfigure}[b]{.5\textwidth}
\centering
  \includegraphics[scale=.14]{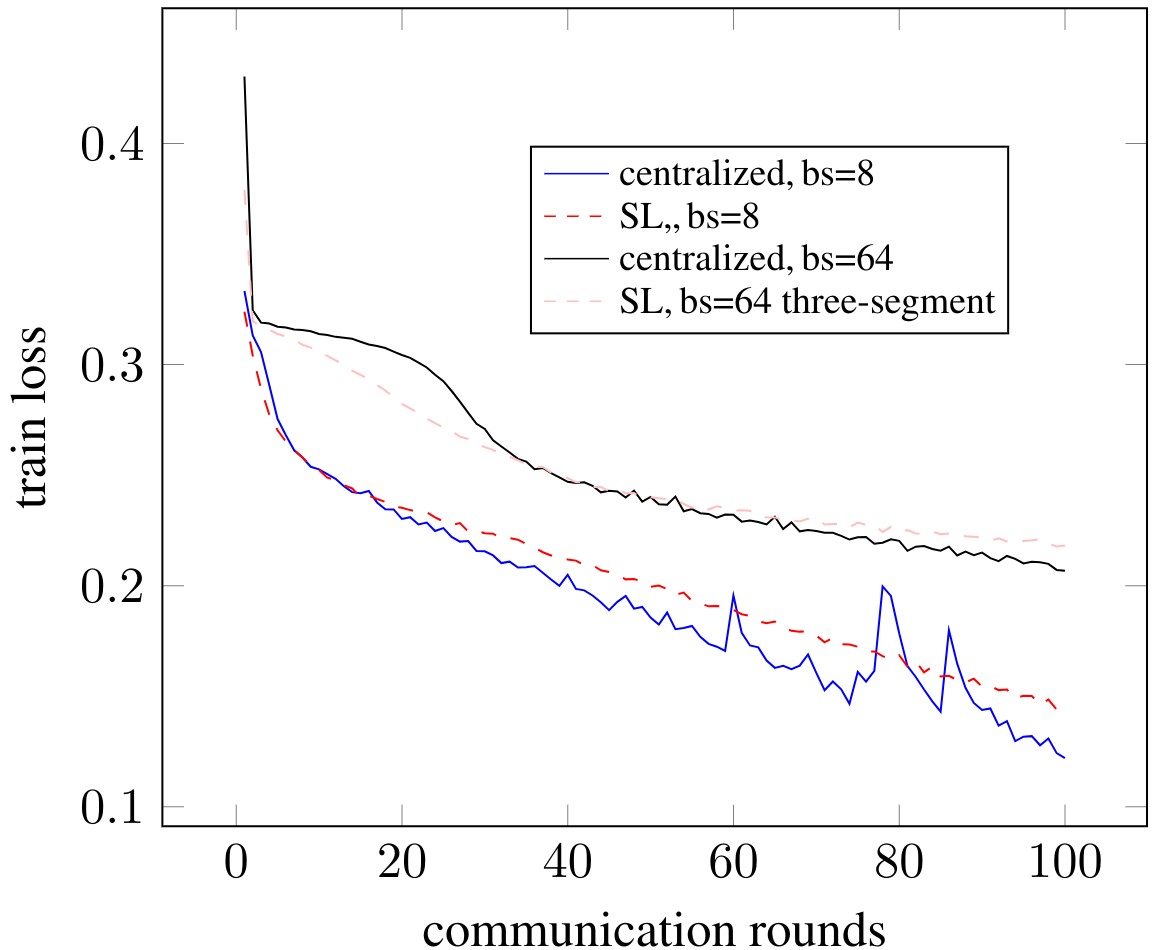}
\end{subfigure}
\begin{subfigure}[b]{.5\textwidth}
\centering
  \includegraphics[scale=.14]{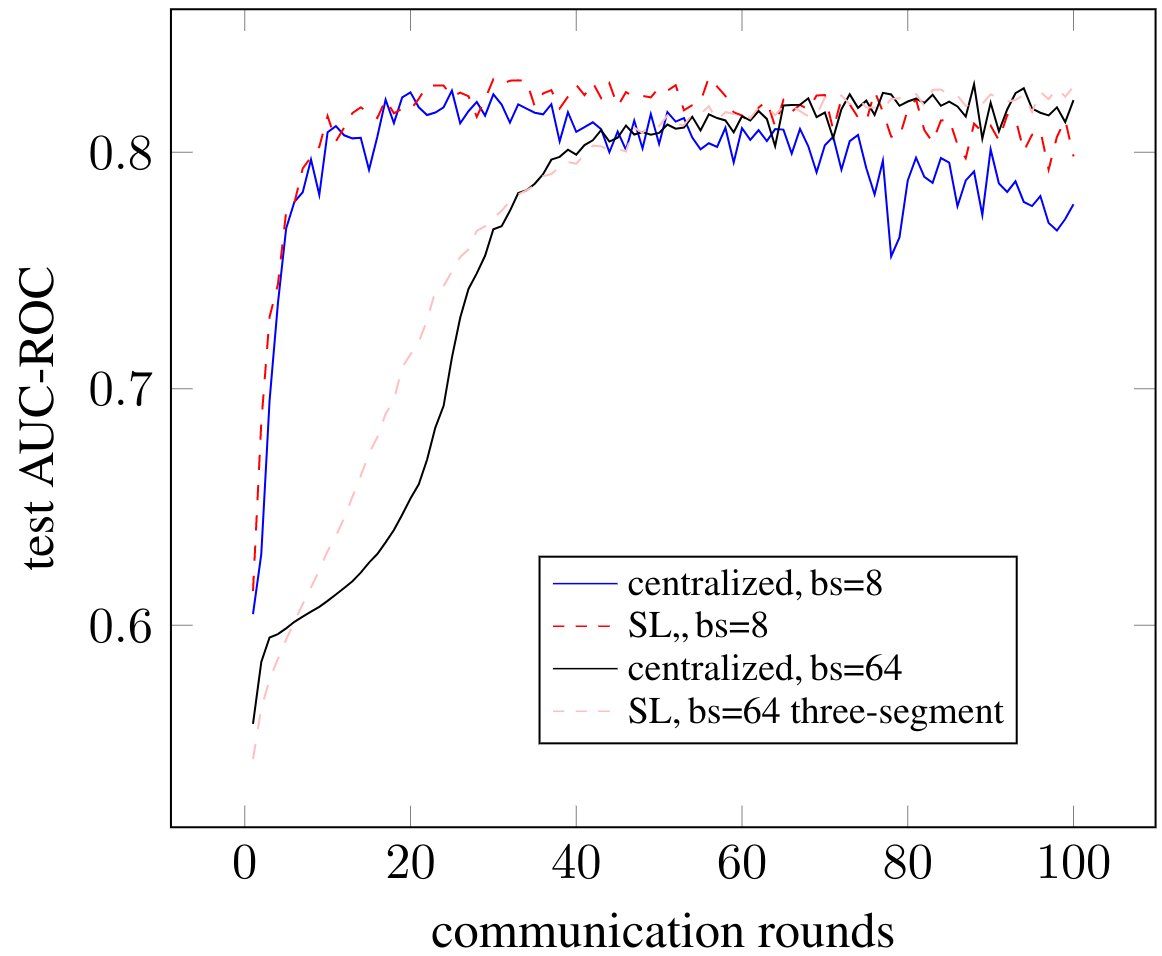}
\end{subfigure}

\caption{Train loss (left) and test AUC-ROC (right) of applying centralized learning and the proposed SL approach to the eICU dataset for different values of batch sizes (bs). The proposed SL method for RNNs successfully follows the centralized learning and achieves higher AUC-ROC, while handling sequentially distributed data and preserving privacy.}
\label{fig:fig12}
\end{figure}

\noindent\textbf{Comparing the proposed SL method for RNNs with centralized learning:} Figure \ref{fig:fig12} shows the train loss and AUC (Area Under Curve) of ROC (Receiver Operating Characteristics) curves versus 100 communication rounds of the proposed SL method and the centralized  learning on the eICU dataset. We performed experiments for batches of size 8 and 64. As can be observed in the learning curves in Figure. \ref{fig:fig12}, in all cases, the proposed SL method for RNNs successfully follows the centralized learning and achieves higher AUC-ROC, while handling sequentially distributed data and preserving privacy.

\begin{figure}[ht]

\begin{subfigure}[b]{.5\textwidth}
\centering
  \includegraphics[scale=.14]{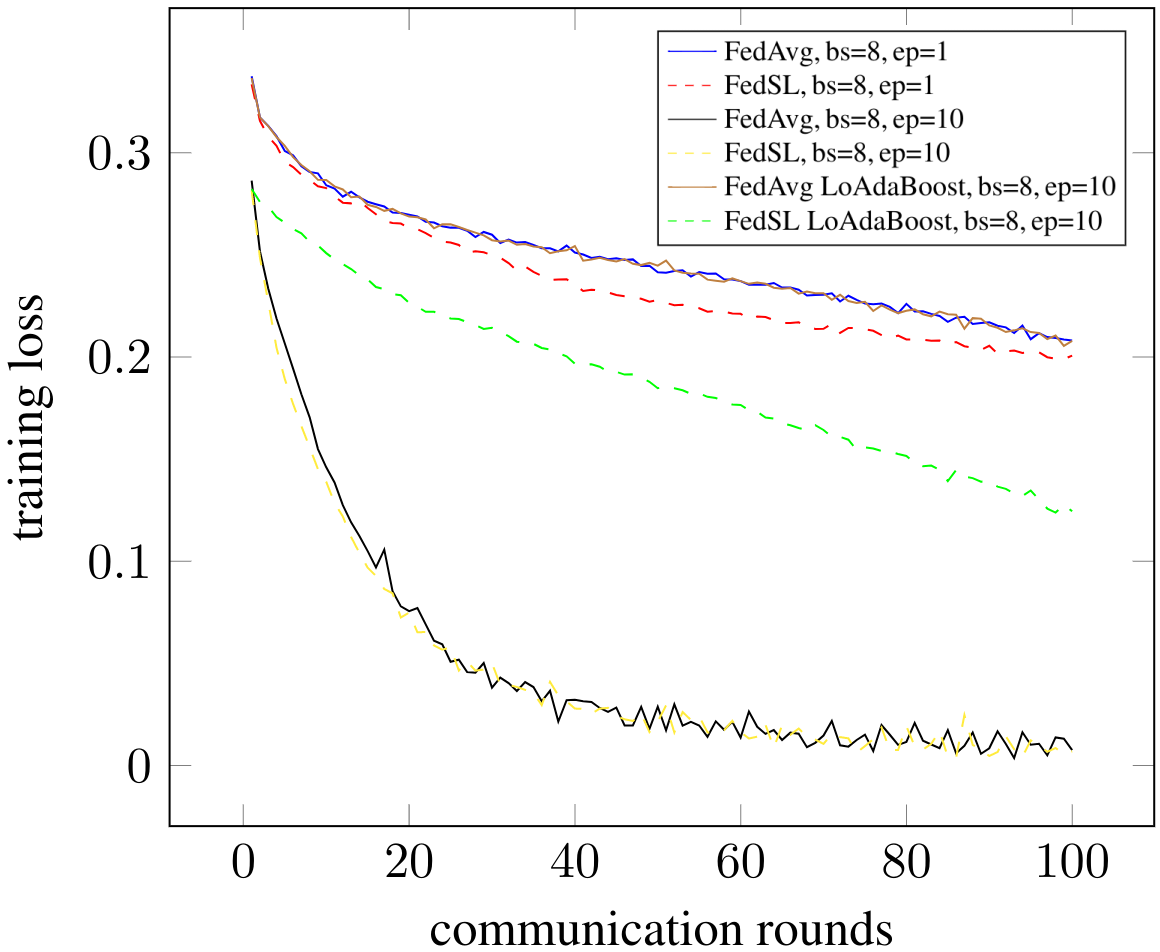}
\end{subfigure}
\begin{subfigure}[b]{.5\textwidth}
\centering
  \includegraphics[scale=.14]{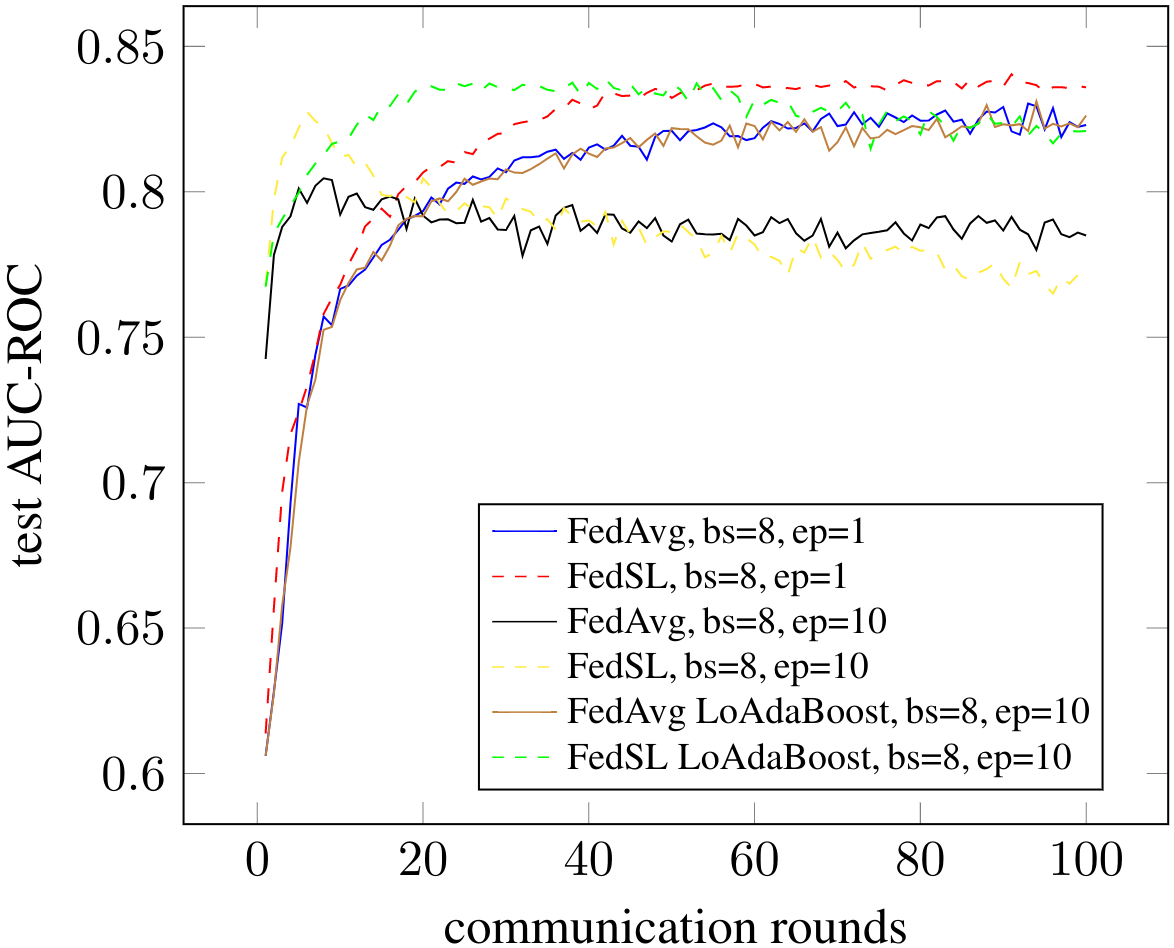}
\end{subfigure}
\begin{subfigure}[b]{.5\textwidth}
\centering
  \includegraphics[scale=.185]{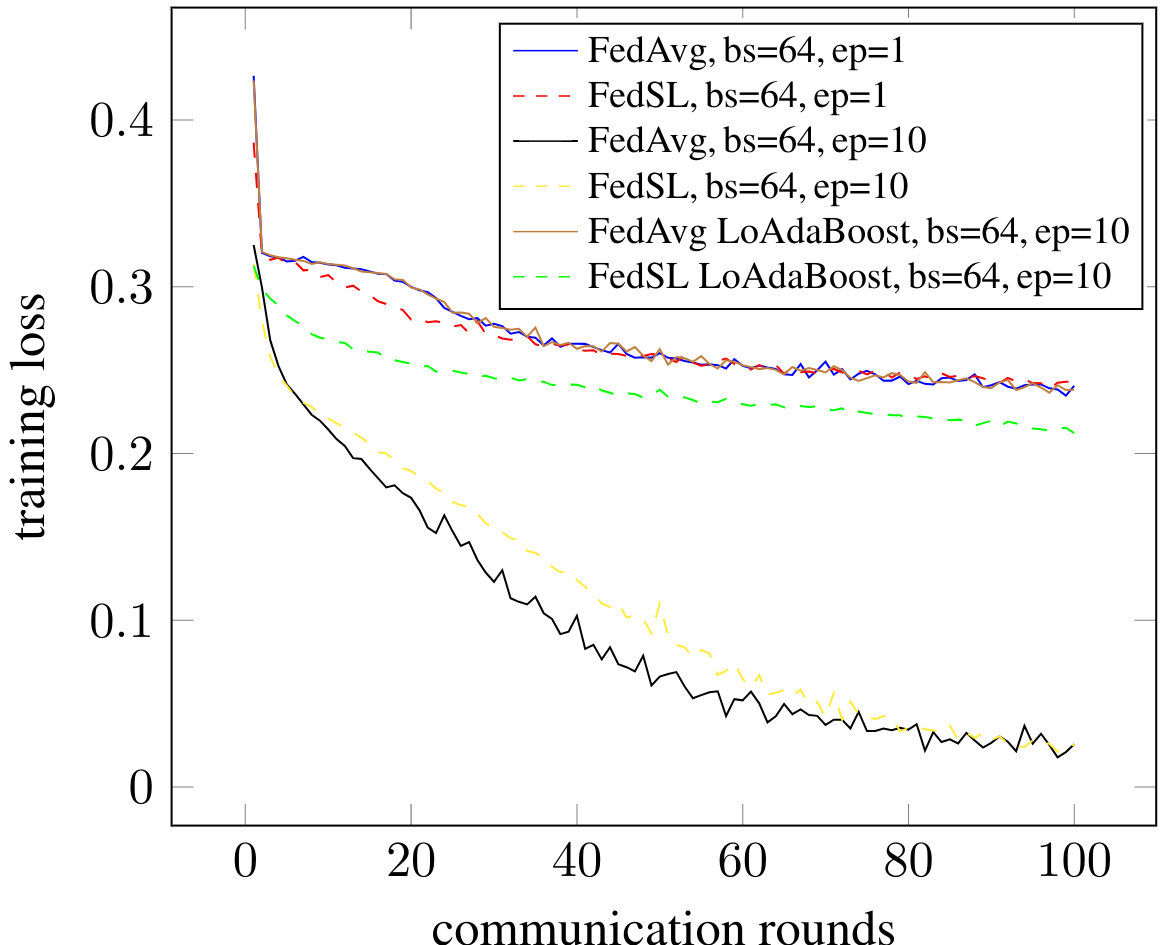}
\end{subfigure}
\begin{subfigure}[b]{.5\textwidth}
\centering
  \includegraphics[scale=.14]{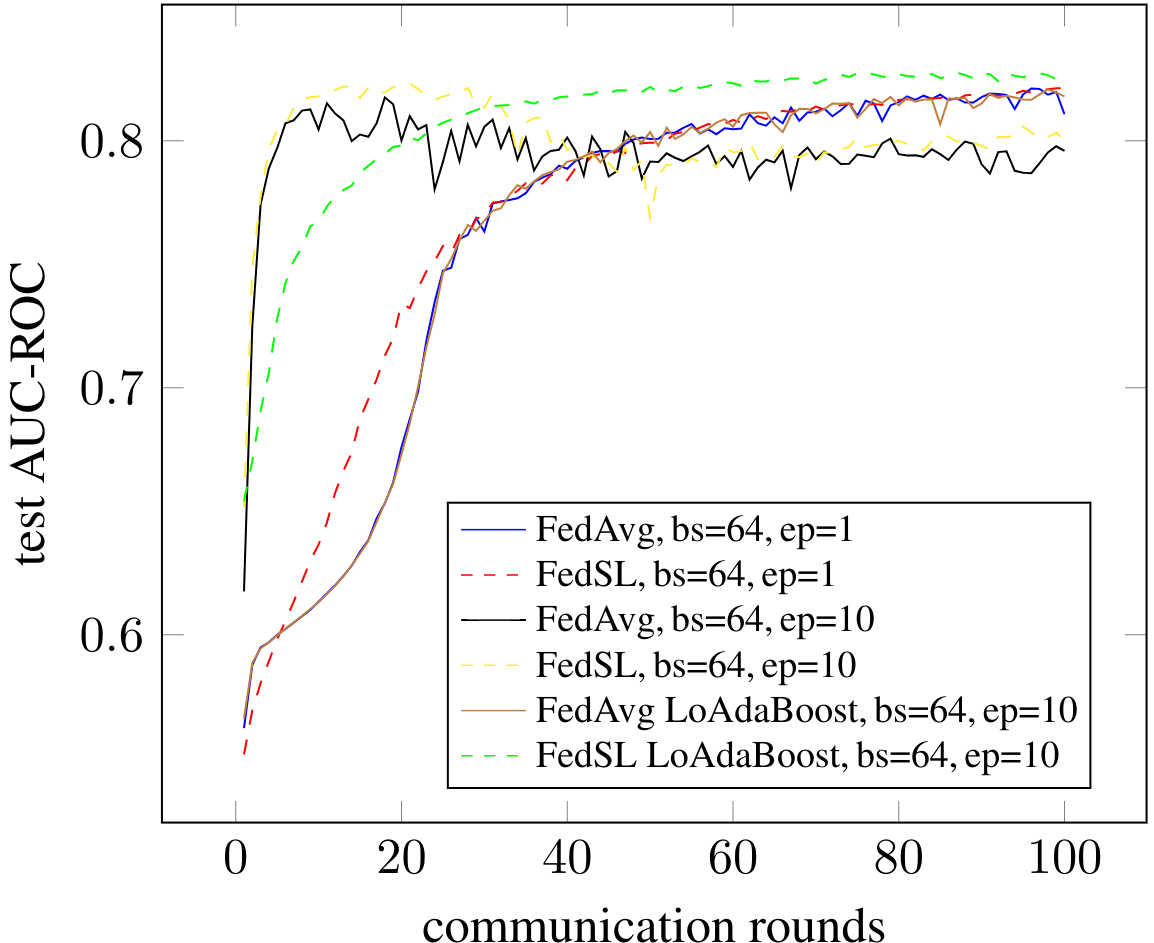}
\end{subfigure}

\caption{Train loss (left column) and test AUC-ROC (right column) of applying different FL approaches to the eICU dataset for different values of batch sizes (bs) and local epochs (ep).}
\label{fig:fig13}
\end{figure}

\noindent\textbf{Local computation per client:} In Figure. \ref{fig:fig13}, we report the results of applying the proposed FedSL to the eICU dataset compared to the FedAvg \cite{mcmahan2017communicationefficient}, LoAdaBoost FedAvg \cite{huang2020loadaboost}, and also the combination of the proposed FedSL with LoAdaBoost FedAvg, for local batches of size 8 and 64, and local epochs of size 1 and 10. As can be observed in Figure. \ref{fig:fig13}, in terms of AUC-ROC, the proposed FedSL outperforms FedAvg, and also the combination of FedSL with LoAdaBoost outperforms LoAdaBoost FedAvg.

\section{Conclusions and Future Work} \label{sec:cf}
In this paper, we defined and addressed the problem of training models in a federated privacy-preserving setting on distributed sequential data occurring frequently in practice. Sequentially partitioned data is different from the horizontal and vertical partitioned data studied in the previous FL and SL works \cite{yin2021comprehensive, thapa2021advancements}. We presented a novel architecture that integrates FL and SL to address the problem of training models on sequentially partitioned data. In this architecture, none of the data, or label, or complete model parameters are shared between clients or between clients and the server. In order to analyze multiple segments of distributed sequential data, we introduced a novel SL approach tailored for RNNs. Consecutive splits of RNNs are trained on consecutive clients containing consecutive segments of multiple-segment sequential data distributed across clients. This SL approach for RNNs is different from the previous SL approaches working on feed-forward neural networks. 

The experimental results on simulated and real-world multiple-segment sequential data (with different values of local epochs and local batch sizes, with different numbers of participants, and on IID and non-IID data) demonstrate that the proposed method successfully trains models on distributed sequential data, while preserving privacy, and outperforms previous FL and centralized learning approaches in terms of achieving higher accuracy in fewer communication rounds.

In future research, we plan to work on split learning in other sequential neural networks, including Temporal Convolutional Networks (TCNs), 3D CNNs, and use the split sequential models in our proposed FedSL framework. In addition, we aim to apply modified versions of FedAvg \cite{he2020fedml} in the FedSL framework. Other directions of our further research are utilizing differential-privacy techniques in the FedSL framework, and working on techniques in SL to personalize split models for individual clients.
\\\\
\textbf{Data availability}\\
The datasets analyzed during the current study are publicly available in the following repositories:\\
http://yann.lecun.com/exdb/mnist/\\
https://github.com/zalandoresearch/fashion-mnist\\
https://eicu-crd.mit.edu/about/eicu/
\\\\
\textbf{Conflict of Interest}\\
The authors declare that they have no conflict of interest.

\bibliography{sn-bibliography}
\end{document}